\pdfoutput=1

\documentclass[11pt]{article}

\usepackage{array}

\newcolumntype{P}[1]{>{\centering\arraybackslash}p{#1}}

\usepackage{acl}

\usepackage{svg}
\usepackage{times}
\usepackage{latexsym}
\usepackage{booktabs}  
\usepackage{graphicx}
\usepackage{multirow}
\usepackage{colortbl}
\usepackage{xspace}
\usepackage[most]{tcolorbox}
\usepackage{enumitem}
\usepackage{subcaption}

\definecolor{distresscolor}{HTML}{ffc9c9}
\definecolor{culturecolor}{HTML}{a5d8ff}
\definecolor{supportcolor}{HTML}{fa5252}

\setlist{noitemsep, topsep=0pt}

\usepackage[T1]{fontenc}

\usepackage[utf8]{inputenc}

\usepackage{microtype}
\usepackage{pifont}
\usepackage{tabularx}
\usepackage{xcolor, colortbl}
\newcommand{\NA}{---}
\definecolor{pagreen}{HTML}{d0e0e3}
\definecolor{pared}{HTML}{ead1dc}
\definecolor{pablue}{HTML}{c9daf8}
\definecolor{padgreen}{HTML}{a2c4c9}
\definecolor{padred}{HTML}{d5a6bd}

\newcommand{\greencell}{\cellcolor{pagreen}}

\newcommand{\emocell}{\cellcolor{distresscolor}}
\newcommand{\cultcell}{\cellcolor{culturecolor}}

\definecolor{customdarkgreen}{RGB}{50,215,50}
\newcommand{\redxmark}{{\color{red}\ding{55}}}
\newcommand{\greencheck}{{\color{customdarkgreen}\ding{51}}} 
 
\newcommand{\methodname}[1]{\texttt{\textbf{CultureCare}}\xspace}

\title{Tailored Emotional LLM-Supporter: Enhancing Cultural Sensitivity}

\author{Chen Cecilia Liu$^{*1}$, Hiba Arnaout$^{*1}$, Nils Kovačić$^1$, Dana Atzil-Slonim$^2$, Iryna Gurevych$^1$ \\
  1. Ubiquitous Knowledge Processing Lab (UKP Lab) \\
  Department of Computer Science and Hessian Center for AI (hessian.AI) \\
  Technische Universität Darmstadt\\
  2. Department of Psychology, Bar-Ilan University}

\begin{document}
\maketitle
 \def\thefootnote{*}\footnotetext{Equal Contributions.}
\def\thefootnote{\arabic{footnote}}
\begin{abstract}
Large language models (LLMs) show promise in offering emotional support and generating empathetic responses for individuals in distress, but their ability to deliver culturally sensitive support remains underexplored due to a lack of resources. In this work, we introduce \methodname{}, the first dataset designed for this task, spanning four cultures and including 1729 distress messages, 1523 cultural signals, and 1041 support strategies with fine-grained emotional and cultural annotations. Leveraging \methodname{}, we (i) develop and test four adaptation strategies for guiding three state-of-the-art LLMs toward culturally sensitive responses; (ii) conduct comprehensive evaluations using LLM-as-a-Judge, in-culture human annotators, and clinical psychologists; (iii) show that adapted LLMs outperform anonymous online peer responses, and that simple cultural role-play is insufficient for cultural sensitivity; and (iv) explore the application of LLMs in clinical training, where experts highlight their potential in fostering cultural competence in novice therapists. Github: \url{https://github.com/UKPLab/eacl2026-culturecare}.\footnote{\textcolor{purple}{Content Warning: This paper includes examples that some readers may find offensive or triggering. These instances are presented for research purposes only and do not represent the views of the authors or affiliated institutions.}}
\end{abstract}

\begin{figure}[t]
    \centering
    \includegraphics[width=0.85\linewidth]{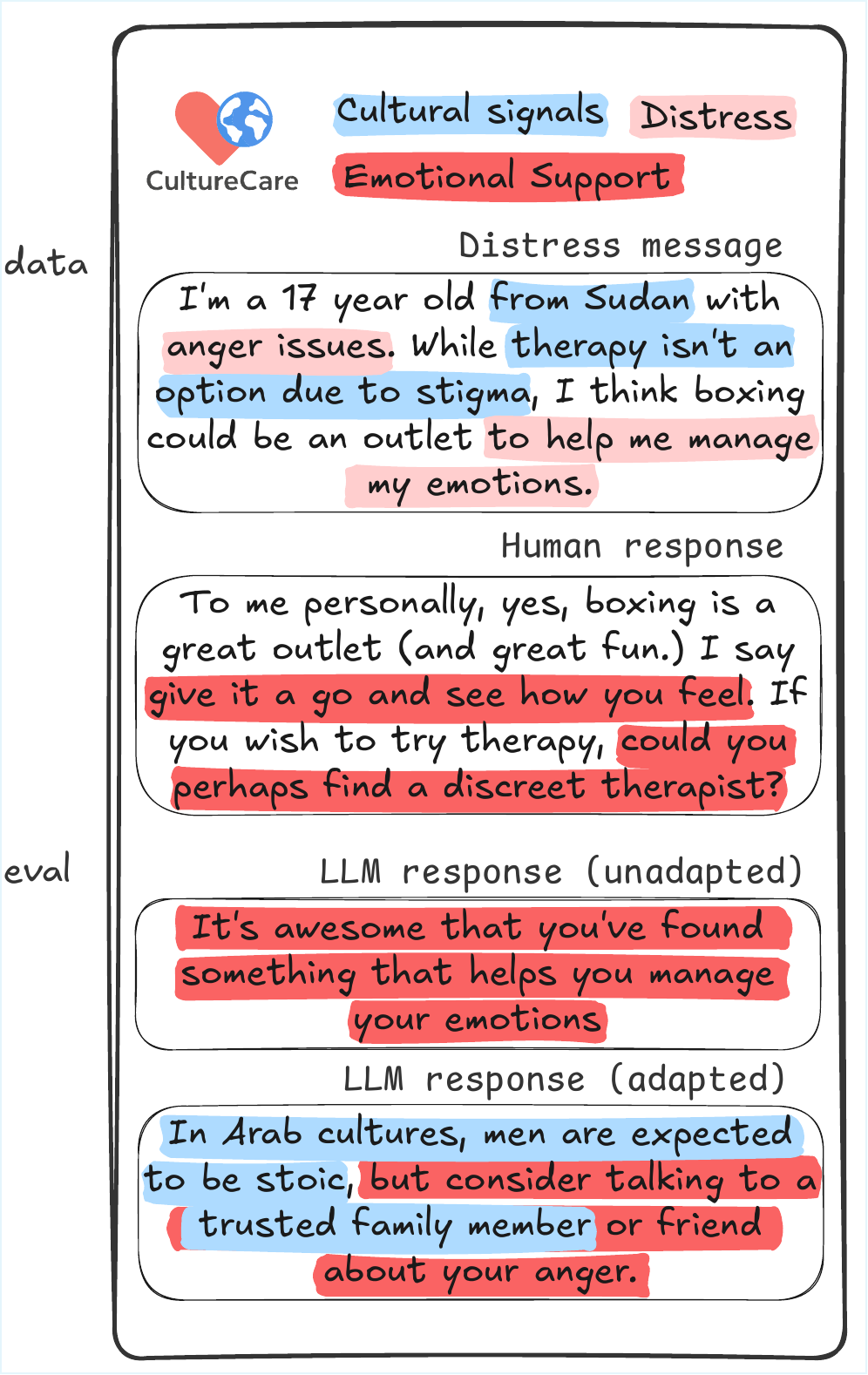}
    \caption{\methodname{}: 1. The ``data'' panel shows a Reddit post span-annotated for emotional distress and cultural signals. Every post is paired with its top Reddit response, span-annotated for emotional support messages and cultural signals. 2. The ``eval'' panel shows the responses to the post returned by LLMs, with and without cultural adaptation, respectively.} 
    \label{fig:figureone}
\end{figure}

\section{Introduction}
Large language models (LLMs) have shown growing potential in offering emotional support, with recent work demonstrating that LLMs can provide empathetic, contextually relevant responses for individuals experiencing distress \cite{acl/zheng-etal-2024-self, zhan2024large, coling/ye-etal-2025-sweetiechat}. This emerging capability is particularly promising in online spaces, where peer support communities play a vital role in helping individuals navigate emotional challenges. However, culture shapes human emotional experiences and influences the stressors people encounter in daily life \cite{markus1991culture, mesquita1992cultural, chun2006culture, mesquita2017doing}. As a result, one critical dimension remains underexplored---the \emph{cultural sensitivity} of these LLMs' support responses.

Effective emotional support is deeply shaped by cultural knowledge, unspoken assumptions, norms, and values that influence how distress is expressed and how support is received \cite{taylor2007cultural, matsumoto2008mapping, kim2008culture}. However, when LLMs are used as tools for emotional support, they may struggle to recognize these culturally embedded cues of distress in the first place \cite{10.1145/3678884.3681858}. Even well-intentioned responses can cause harm or alienation without cultural grounding and sensitivity, rather than providing the support users seek \cite{lissak-etal-2024-colorful, moore2025expressing}.  For example, in a collectivist culture, an LLM advising a user to cut ties with their family for personal happiness, without acknowledging the cultural weight of familial duty, may seem offensive or immoral. Similarly, in cultures where mental health is heavily stigmatized, bluntly urging someone to ``seek therapy'' might intensify feelings of shame or social isolation rather than offering relief.

Recent research tries to address this issue through a case study on Pakistani culture \cite{10.1145/3678884.3681858}. However, the study offers limited generalizability due to its narrow set of manually created distress scenarios, leaving the broader challenge of culturally sensitive emotional support largely unexplored. The lack of suitable datasets and the evaluation of adaptation methods have hindered progress in this area. Therefore, here, we present a comprehensive multi-cultural investigation into LLMs' ability to provide culturally sensitive emotional support.

To address the data gap, we introduce \methodname{}. To the best of our knowledge, this is the first dataset designed to support the study of culturally-sensitive peer emotional support (Figure \ref{fig:figureone}). \methodname{} spans four distinct global cultures, namely Arabic, Chinese, German, and Jewish, and collected fine-grained annotations for both emotional support strategies and culturally-relevant signals, as shown in Table~\ref{tab:comparisons}. The dataset consists of distress-response pairs sourced from real-world interactions from Reddit, annotated for support type and cultural signals, allowing for nuanced development and evaluations for adapting LLMs' responses across cultural contexts. Using \methodname{}, we evaluate three state-of-the-art LLMs with tailored prompting strategies for adaptation. Through both automated evaluations and human evaluations, we find that while incorporating basic cultural information helps, a more effective adaptation requires detailed guidelines and attention to contextualized, explicit cultural signals. 

While our primary focus is on peer support---where the LLM acts as a supportive peer rather than providing professional help---we also explore its potential in clinical training settings of training psychology students to conduct culturally competent therapy~\cite{benuto2018training}. Expert feedback highlights strong safety and promising utility.

To sum up, our contributions are: First, we release the first dataset, \methodname{}, for evaluating and adapting LLMs in culturally-sensitive emotional support, spanning four cultures with fine-grained annotations. The dataset comprises 1729 annotated distress messages, 1523 cultural signals, and 1041 support strategies; second, we develop and test four \textbf{adaptation strategies} to guide three popular state-of-the-art LLMs toward generating culturally-sensitive support responses; and third, we provide \textbf{comprehensive evaluations} involving LLM-as-a-Judge, in-culture human evaluators, and clinical psychologists to assess both the emotional and cultural aspects of the generated responses.

\begin{table*}[ht]
\centering
\resizebox{0.95\linewidth}{!}{
\begin{tabular}{l c p{2.8cm} l p{3cm} l}
\toprule
 & \textbf{Real?} & \textbf{Cultures} & \textbf{Size} & \textbf{Annotations} & \textbf{Eval. Aspects}  \\
\hline
\citet[ESConv]{liu2021towards} & \greencheck  & \NA  &1053  & DM, SS, E  & E \\
\citet[ExTES]{acl/zheng-etal-2024-self} & \redxmark & \NA   & 11177$^\dagger$ & DM, SS  & E\\
\citet[FEEL]{zhang2024feel} & Mix & \NA  & 200 &  DM, SS & E \\
\citet{10.1145/3678884.3681858} & \redxmark & Pakistani & 7 & \NA & C \\
\hline
\cellcolor{gray!15}\methodname{} & \cellcolor{gray!15}\greencheck & \cellcolor{gray!15}Arabic, Chinese, German, Jewish & \cellcolor{gray!15}462 & \cellcolor{gray!15}DM, SS, Cultural Signals & \cellcolor{gray!15}E, C \\
\bottomrule
\end{tabular}
}
\caption{Comparison with existing work on LLMs for emotional support. \textbf{E}: Emotion, \textbf{C}: Culture, \textbf{DM}: Distress messages, \textbf{SS}: Support strategies. $\dagger$: \citet{acl/zheng-etal-2024-self} contains a subset of 101 dialogues on cultural identity and belonging; however, culture is not a focus of their work. \methodname{} uniquely focuses on \emph{culture}, explicitly annotates spans which include cultural signals and assigns their types, and evaluates emotional support responses from both emotional and cultural perspectives. Our dataset comprises 4,293 annotations, as detailed in Table~\ref{tab:stats}.}
\label{tab:comparisons}
\end{table*}

\section{Related Work}

\noindent\textbf{Culturally adapted LLMs.}
Recent research has found that LLMs predominantly reflect the perspectives of WEIRD (Western, Educated, Industrialized, Rich, and Democratic, \citealt{henrich2010weirdest}) populations without any adaptation \cite{atari2023humans, ghost_has_american_accent}. Several studies attempted to address this issue, focusing on diverse tasks such as value alignment \cite{nvm/anthroprompt/acl/abs-2402-13231, acl/clca} or hate speech classification \cite{hate/eacl/zhou-etal-2023-cross, cultureLLM/corr/abs-2402-10946, adilazuarda2025surveys}. \citet{nvm/anthroprompt/acl/abs-2402-13231, pnas/tao2024cultural} demonstrated that prompting LLMs with cultural and persona-specific information can effectively align models with diverse cultural values. However, existing work has not examined the effectiveness of these prompting methods for culturally aware emotional support---an important gap this study addresses.

\noindent\textbf{Culture and mental health.} Culturally sensitive counselling is a well-established consideration in clinical psychology and healthcare settings~\cite[among others]{evm/bernal1995ecological, cp/csm/42269711-6e5b-382a-b370-098ad7128da0, kreuter2004role/culture_in_health, taylor2007cultural, tao2015meta}. Prior research has explored various aspects of incorporating cultural sensitivity in practical domains outside of NLP, including the importance of cultural humility in improving therapy outcomes~\cite{owen2016client}, disparities in engagement and follow-up care across demographics~\cite{zeber2017impact}, and the need to embed cultural competence in training programs~\cite{benuto2018training}. However, the application of LLMs in culturally sensitive mental health remains limited. Focusing on formal therapy settings, recent work~\cite{abbasi2025hamrazculturebasedpersianconversation, kim2025kmidatasetkoreanmotivational} explores LLM-generated synthetic clinical conversations in multilingual contexts. Their focus on clinical therapy and synthetic data generation differs from ours, which centers on examining LLMs for culturally-aware emotional support across several LLMs and adaptation strategies.

\noindent\textbf{LLMs for emotional support.} Existing work has shown that LLMs can provide empathetic and supportive responses when appropriately guided \cite{zhan2024large}. Studies such as \citet{liu2021towards, acl/zheng-etal-2024-self, zhang2024feel} examine how LLMs respond to distress messages using various support strategies. However, they largely overlook the influence of cultural context in shaping emotional needs and support preferences. While \citet{10.1145/3678884.3681858} considers cultural context, its focus on a small set of scenarios from the Pakistani culture limits its generalizability. Furthermore, recent studies on LLMs in cognitive behavioural therapy \cite{naacl/goel-etal-2025-socratic, abs-2504-17238/cbt, zhang-etal-2025-cbt} emphasize clinical settings with structured treatment frameworks, offering little attention to cultural nuance. In contrast, our work focuses on emotional support where cultural understanding is \textit{essential} for effective and supportive communication. We address this gap by incorporating cultural signals from diverse cultural communities into our annotation process, and by assessing emotional support responses from both humans (Reddit users) and LLMs, based not only on evaluation metrics like empathy, but also on cultural metrics like showing an understanding of the cultural context.

\section{\methodname{}} 
\label{sec:dataset}
\begin{figure*}
    \centering
    \includegraphics[width=0.9\linewidth]{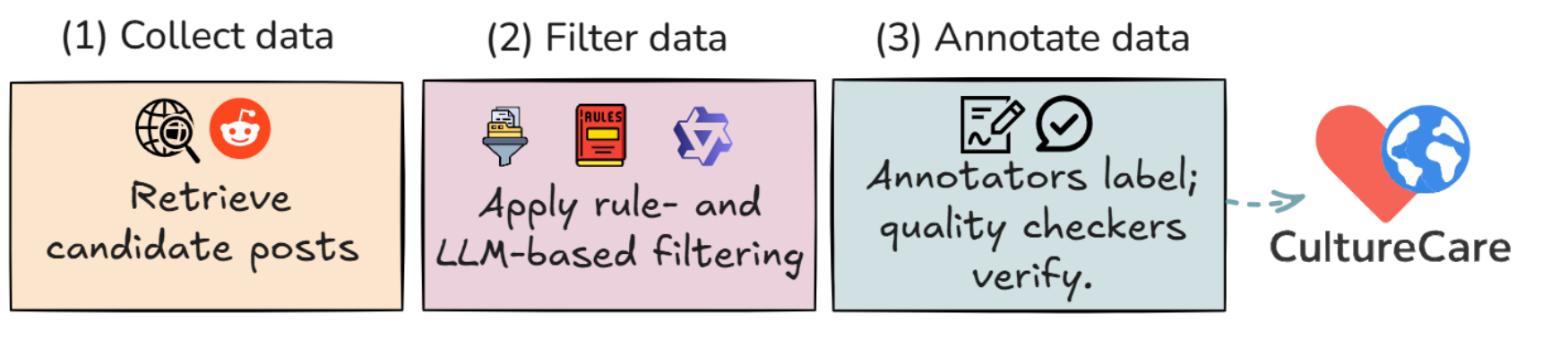}
    \caption{The \methodname{} dataset construction pipeline: (1) we collect data by querying selected subreddits for candidate post-response pairs; (2) we apply rule-based and LLM-based filters to remove noisy instances (\S\ref{subsec:filter}), e.g., that do not contain cultural signals; (3) in-culture annotators mark spans, in both posts and responses, with emotional distress, cultural signals, and support strategies; finally, a second group of annotators verify the quality of these labels and make corrections when needed.}
    \label{fig:creationpipeline}
\end{figure*}

We present \methodname{}, a multi-cultural dataset with fine-grained span-level annotation of cultural signals and emotional distress.\footnote{We deliberately chose not to use language as the defining boundary of culture, recognizing that culturally influenced distress can be expressed in any language in online communities. As a result, \methodname{} includes post-responses pairs both in English and in the native languages associated with each culture. The language distribution varies by culture; see Appendix~\ref{app:language}.} Briefly, we begin by collecting publicly available Reddit\footnote{An anonymous platform where user identities remain hidden.} posts at the intersection of culture and mental health. Specifically, we draw from mental health subreddits using culture-related keywords (e.g., \texttt{r/depression} with the keyword ``\textit{Chinese}'') and vice versa (e.g., \texttt{r/china} with the keyword ``\textit{depression}''). We then apply a combination of rule-based and LLM-based filters to remove noise (i.e., irrelevant posts for the task we target). Every instance consists of a post-response pair. Finally, in-culture annotators annotate these instances along the following dimensions: emotional distress and their intensity, cultural signals and their categories, support messages and their strategies, and overall empathy level. This is followed by quality checks by in-culture reviewers. To protect annotators' mental health, we provide clear content warnings for sensitive topics and allow annotators to stop the task at any time. The construction pipeline is illustrated in Figure~\ref{fig:creationpipeline}. This research was approved by the ethics committee of the Technical University of Darmstadt (EK 121/2024).


\begin{table}
\centering
\resizebox{0.99\linewidth}{!}{
\begin{tabular}{l l l l l l}
\toprule
  \textbf{Category} & \textbf{AR} & \textbf{CH} & \textbf{GE} & \textbf{JE} & \textbf{All}  \\
\hline
$\#$ posts & 110 & 141 & 119 & 92 & 462\\ 
$\#$ responses & 104 & 126 & 100 & 88 & 418\\ 
$\#$ distress messages & 397 & 399 & 402 & 531 & 1729\\ 
$\#$ cultural signals & 346 & 315 & 338 & 524 & 1523\\ 
$\#$ support strategies  & 259 & 242 & 194 & 346 & 1041\\ 
$\#$ demographic info & 226 & 301 & 268 & 131 & 926\\ 
Avg. post length & 316 & 492 & 690 & 389 & 480\\ 
Avg. response length & 101 & 80 & 125 & 88 & 98\\ 
\bottomrule
\end{tabular}
}
\caption{\methodname{} statistics. \textbf{AR}: Arabic, \textbf{CH}: Chinese, \textbf{GE}: German, and \textbf{JE}: Jewish. }
\label{tab:stats}
\end{table}

\subsection{Data Collection and Filtering}\label{subsec:filter}
\noindent \textbf{Candidate posts.} We use Reddit as our dataset source due to its global user base and peer-driven mental health discussions. Our data is collected via the Reddit API\footnote{\url{https://www.reddit.com/dev/api/}}. We focus on two setups: 1. searching mental health subreddits with culture-specific keywords, and 2. searching culture-specific subreddits with mental health keywords. The list of subreddits and keywords is in Appendix~\ref{datacollection}. For each relevant post, we fetch the top-voted comment as the ideal supportive response, as it typically offers emotional support and resonates with readers, reflected in its high upvotes. 

\noindent \textbf{Filtering.} Our initial dataset contained 9160 posts, many of which were noisy---e.g., general mental health tips, reactions to global events, or content unrelated to the target cultures. Since manual review of all posts was infeasible, we first applied rule-based filters to remove both explicit noise (e.g., URLs-only posts) and LLM-based filtering to remove implicit noise (e.g., posts lacking \textit{personal} distress).  This leaves us with 2671 posts for manual review. The full set of filters is detailed in Appendix~\ref{filtering}. After the final manual review, where annotators were asked to flag any remaining irrelevant posts and not annotate it (e.g., \textit{a post from someone who lives in Germany on one of the German subreddits, but is not culturally German}), we retained 462 high-quality instances. While small, this dataset allows for focused, fine-grained annotation, and is the largest of its kind (more discussions in Appendix \ref{sec:462}).

\subsection{Annotations and Quality Assurance}
Each post-response was annotated by an in-culture annotator, who selects the spans and labels the data for the following dimensions: \textbf{emotional distress} (to highlight the span of text where the poster expresses emotional distress), \textbf{intensity of emotional distress}  (how intense are the emotions in the post), \textbf{cultural signals} (to highlight any text indicative of a culture reference in both the post and the response), \textbf{types of cultural signals} (to categorize the annotated cultural signal, e.g., values), \textbf{support messages} (to highlight the span of text in the response where the responder offers support), the \textbf{support strategy} behind the message (to categorize the annotated support message, e.g., offering suggestions) and its \textbf{empathy level} (how empathetic is the support message in the response). To ensure high-quality annotations, an in-culture quality checker reviewed the initial annotations and left comments when they disagreed; the annotator then reviewed and approved these comments. Most additions were suggestions for extra annotations rather than indications of inconsistencies. On average, each post took approximately 10 minutes to annotate and 5-7 minutes to review, as they were often lengthy and required attentive reading. 

The definitions and categories for these dimensions are detailed in Appendix \ref{app:anno_guidelines}, along with guidelines and measures for content sensitivity.

\subsection{Analysis of \methodname{}}
\noindent \textbf{Statistics.} \methodname{} includes 462 posts\footnote{90\% with responses. The remaining 10\% had deleted comments or no comments.}, containing 1729 annotated distress messages, 1523 cultural signals, and 1041 support strategies. We also extract demographic details (e.g., age, gender, religion) from the posts using an LLM, yielding 926 additional annotations. An overview of these statistics per culture is in Table~\ref{tab:stats}, and sample annotated posts are in Appendix~\ref{app:examples}.

\noindent 
\textbf{Prevalent cultural signals and support strategies by culture.} In order to understand the most frequent culture category and emotional support strategy, we compute the number of occurrences of the category, i.e., type, of every culture signal (namely, concepts, knowledge, values, norms and morals, language slang, artifacts, and demographics; definitions and examples in Table~\ref{tab:cultureschema}), and every support message strategy (questions, restatement, reflection of feelings, self-disclosure, affirmation, suggestions, information; definitions and examples in Table~\ref{tab:supportchema}), annotated by in-culture annotators and approved by quality checkers.  We found that the most recurring cultural signals are as follows: Arabic (\textit{values}), Chinese (\textit{norms and morals}), German (\textit{norms and morals}), and Jewish (\textit{concepts}). Moreover, the most recurring support strategy is ``\textit{providing suggestions}'' for \textit{all} cultures. The full distributions of these categories are in Figure~\ref{fig:categoryperculture} in Appendix~\ref{app:categoryperculture}. 

\noindent 
\textbf{Demographic diversity.} Using GPT-4o-mini, we extract (\textit{when present}) detailed demographic information~\footnote{This goes beyond the culture-related demographic information in our human annotations.}, namely place of residence, gender, age, born in, marital status, number of people in the household, education, profession, employment, class, and religion. Every extraction comes with both an answer and evidence. For example, out of the evidence (span of text) ``\textit{I'm a 17M living in Sudan}'', the model extracts \textit{gender:male, age:17}. When any of the demographic fields is not found, the LLM returns ``unknown'' for that field. We manually inspected a subset of the extractions and found no evidence of inconsistencies or hallucinations. For 50\% of the dataset, demographic information could be extracted. We release these LLM-derived annotations alongside the dataset.  Finally, we provide a qualitative analysis of the demographic diversity in \methodname{} across cultures. The dataset encompasses a broad range of cultural backgrounds and life circumstances, including diverse geographic origins (e.g., Arabic: from Syria, Egypt, Saudi Arabia, and more), age groups (e.g., Chinese: ranging from 15 to 58 years), and professions (e.g., German: working as pizza couriers, social workers, opticians, etc.). This reflects substantial cultural and social heterogeneity within each group. Additional details are in Table~\ref{tab:demoinfo}, Appendix~\ref{app:demographicinfo}, and the released dataset.

\noindent 
\textbf{Prevalent norms, morals, and values.} 
In this analysis, we focus on the cultural signals human-annotated in the posts and responses that were categorized, by the in-culture annotators, specifically under ``\textit{norms and morals}'' and ``\textit{values}''.  To understand culture-specific themes under this category, we prompt GPT-4o, to cluster, per culture, the spans annotated as norms and values, and return the top 5 clusters (by descending order of size). We manually inspect the resulting clusters and find no inconsistencies. The top clusters per culture are shown, with examples for every cluster, in Table~\ref{tab:normsandvalues}. This analysis shows that emotional struggles are universal, but their expression and underlying causes vary across cultures, revealing both shared and distinct themes.

\begin{table*}
\centering
\resizebox{0.85\linewidth}{!}{
\begin{tabular}{l l}
\toprule
\textbf{Culture} & \textbf{Themes} \\
\midrule
\textbf{Arabic} & 
\textbf{Mental Health Invalidation:} \textit{mental health is just seen as a phase} \\
& \textbf{Religion Over Mental Health:} \textit{...all they say is I feel this way because I don't pray} \\
& \textbf{LGBTQ+ Rejection:} \textit{came out ... as gay ... I was not met with acceptance} \\
& \textbf{Gender Roles:} \textit{...expects a woman to depend fully on a man and her family}\\
& \textbf{Strict Parenting:} \textit{my dad is extremely strict} \\
\midrule
\textbf{Chinese} & 
 \textbf{Mental Health Invalidation:} \textit{Depression doesn't exist} \\
& \textbf{Verbal Abuse:} \textit{...emotionally abuse me ... how much of a failure and mistake I am} \\
& \textbf{Gender and Identity Issues:} \textit{Asian men are seen as non-masculine ... weak} \\
& \textbf{Family Dynamics:} \textit{It feels like she only lives her life for me} \\
& \textbf{Strict Parenting:} \textit{I live with my tiger parents} \\
\midrule
\textbf{German} & 
\textbf{Relationships and Emotional Dynamics:} \textit{my partner needs time to process emotions} \\
& \textbf{Mental Health and Coping:} \textit{I feel exhausted from pretending to be okay} \\
& \textbf{Family and Childhood Trauma:} \textit{my parents divorced when I was young ... it broke me}  \\
& \textbf{Social Isolation:} \textit{drinking is the only way I sometimes connect socially} \\
& \textbf{Financial Stress:} \textit{I'm on social welfare and feel ashamed} \\
\midrule
\textbf{Jewish} & 
\textbf{Community and Social Stigma:} \textit{In my community a broken engagement is ... a major embarrassment} \\
& \textbf{Personal Spiritual Practice:} \textit{ask god to help those I love and the people around me} \\
& \textbf{Conversion and Identity:} \textit{I have a long history of being told the importance of exact halachic adherence} \\
& \textbf{Gender Roles:} \textit{...men have more responsibilities ... may not be ... much fun} \\
& \textbf{Religion and Mental Health:} \textit{I've started going ... synagogue ... meet with the rabbi for a 1-on-1 session} \\
\bottomrule
\end{tabular}
}
\caption{The five most common themes related to norms and values in \methodname{} are presented in the format (\textbf{theme}: \textit{example}). These themes are expressed by individuals experiencing emotional distress and reflect perspectives rooted in their cultural backgrounds. They do not represent the views or positions of the authors.}
\label{tab:normsandvalues}
\end{table*}

\noindent 
\textbf{Intensity and empathy scores.} Based on the human-annotated intensity level of distress messages in the posts (1: light, 2: moderate, 3: high; definitions in Table~\ref{tab:intensity}) and empathy level of support messages in the responses (from 1:not empathetic at all to 5:very empathetic; definitions in Table~\ref{tab:empathy}), we compute the average intensity and empathy scores across and per cultures. Our findings show that overall, the average intensity level of the distress messages in our dataset is 1.89, per culture: Arabic (1.81), Chinese (1.77), German (1.98), and Jewish (1.89). The average empathy level overall is 2.77, per culture: Arabic (3.27), Chinese (2.15), German (2.73), and Jewish (3.18). 
Since these labels were assigned by in-culture annotators (e.g., Arabic annotators labeling Arabic data), they may reflect cultural bias; future work should include out-of-culture annotators to broaden perspectives.

\section{Adaptation Methods}
\label{sec:adaptation}

We examine three core prompt-based cultural adaptation strategies, namely role-playing, guided principles, and explicit cultural signals, and a combined approach that integrates all of them, alongside a standard Redditor baseline for comparison.

\noindent\textbf{Standard} \texttt{(redditor)}. By default, we prompt the model to be a Redditor, matching the context of the data. This variation serves as a baseline for comparing adaptation strategies. 

\noindent\textbf{Culture-informed role-playing} \texttt{(+culture)}. Building on prior research \cite{nvm/anthroprompt/acl/abs-2402-13231, pnas/tao2024cultural}, instructing LLM to role-play the cultural background of the person is a simple yet effective method for aligning LLM responses with culturally relevant values. Hence, this could enable more empathetic, appropriate responses, removing the \emph{cultural difference} barrier in empathy \cite{cikara2014their, davis2018empathy}.

\noindent\textbf{Guided principles / constitutions} \texttt{(+guided)}. Here, we provide guidelines based on CCCI-R (Cross-Cultural Counselling Inventory—Revised; \citealt{cccir, cccir-dev}, see Appendix \ref{app:cccir} for details), one of the widely established cross-cultural counselling competency measurements by APA (American Psychological Association). This approach aims to emulate some fundamental competency of professional counselling in terms of cultural sensitivity and awareness in an \emph{implicit cross-cultural} setting.

\noindent\textbf{Explicit cultural signals} \texttt{(+annotation)}. We explicitly add the data annotations of posts from our dataset to the prompt. The goal here is to understand whether explicitly providing LLMs with richer contextual information can improve the response in an \emph{implicit cross-cultural} setting.

\noindent\textbf{Combined} \texttt{(+cga)}. In this method, we combine the above three basic strategies, \texttt{\textbf{c}ulture}, \texttt{\textbf{g}uided}, and \texttt{\textbf{a}nnotation}. We modified the guidelines in the \texttt{+guided} strategy by removing CCCI-R items that focus on cross-cultural differences. Here, the LLM will be provided with explicit information and guidelines, as well as role-playing a person from the same culture. 

All adaptation prompts are in Appendix \ref{app:prompts}.

\section{Experimental Setup}
\label{sec:setup}

In this work, we focus on open-source LLMs, prioritizing models in the 7B–8B parameter range due to their strong performance and practical deployment cost, making them well-suited for agentic systems. Our primary evaluation includes Llama-3.1-8B \cite{llm/llama/abs-2302-13971, llama3/corr/abs-2407-21783}, Qwen-2.5-7B \cite{qwen2.5}, and Aya-Expanse-8B \cite{dang2024ayaexpansecombiningresearch}. To examine the robustness of our findings, we additionally test the larger variants of these models. We emphasize open-source models to ensure scientific reproducibility. We use default configurations for generation, and all the adaptation strategies are implemented as system prompts.

\begin{table*}[th]
\centering
\resizebox{0.85\linewidth}{!}{
\begin{tabular}{l ccc ccc ccc ccc ccc}
\hline
\textbf{Model} & \multicolumn{2}{c}{\textbf{Arabic}} & \multicolumn{2}{c}{\textbf{Chinese}} & \multicolumn{2}{c}{\textbf{German}} & \multicolumn{2}{c}{\textbf{Jewish}} & \multicolumn{3}{c}{\textbf{Average}} \\
 & Emo. & Cult.  & Emo. & Cult.  & Emo. & Cult. & Emo. & Cult.  & Emo. & Cult. & All \\
\hline
Aya-Expanse-8B&	 & &  &  &  &  &  & & &  & \\
 \texttt{redditor}	&	4.67 & 3.51 & 4.43 & 3.46 & 4.62 & 2.54 & 4.61 & 3.99 & 4.58 & 3.37 & 3.98 \\
\texttt{+culture}	&	4.55 & 3.56 & 4.48 & 3.63 & 4.61 & 2.75 & 4.67 & 4.12 & 4.58 & 3.51 & 4.05 \\
\texttt{+guided}	&	4.75 & 3.80 & 4.73 & 4.08 & 4.77 & 2.72 & 4.79 & 4.16 & 4.76 & 3.69 & 4.23 \\
\texttt{+annotation}	&	4.77 & 3.73 & 4.77 & 3.78 & 4.80 & 2.75 & 4.81 & 4.10 & 4.79 & 3.59 & 4.19 \\
\texttt{+cga}	&	\emocell4.84 & \cultcell4.39 & \emocell4.82 & \cultcell4.25 & \emocell4.84 & \cultcell3.44 & \emocell4.91 & \cultcell4.55 & \emocell4.85 & \cultcell4.16 & \textbf{4.51} \\
\hline
Qwen-2.5-7B	&	 & &  &  &  &  &  & & &  & \\	
 \texttt{redditor}	&	4.16 & 3.02 & 4.26 & 3.20 & 3.89 & 2.66 & 4.27 & 3.60 & 4.15 & 3.12 & 3.63 \\
\texttt{+culture}	&	4.05 & 3.23 & 4.05 & 3.38 & 3.91 & 2.73 & 4.28 & 3.71 & 4.07 & 3.26 & 3.67 \\
\texttt{+guided}	&	\emocell4.41 & 3.29 & \emocell4.49 & 3.42 & \emocell4.32 & 2.78 & \emocell4.54 & 3.73 & \emocell4.44 & 3.30 & \textbf{3.87} \\
\texttt{+annotation}	&	4.40 & 3.40 & 4.28 & 3.46 & 4.29 & \cultcell2.82 & 4.46 & 3.69 & 4.36 & 3.34 & 3.85 \\
\texttt{+cga}	&	4.11 & \cultcell3.70 & 4.24 & \cultcell3.81 & 3.86 & 2.67 & 4.50 & \cultcell3.95 & 4.18 & \cultcell3.53 & 3.85 \\
\hline
Llama-3.1-8B&	 & &  &  &  &  &  & & &  & \\
 \texttt{redditor}	&	3.79 & 2.98 & 4.20 & 3.41 & 3.81 & \cultcell2.67 & 4.22 & 3.89 & 4.00 & 3.24 & 3.62 \\
\texttt{+culture}	&	3.75 & 3.65 & 4.11 & 3.99 & 3.74 & 2.62 & 4.34 & 4.15 & 3.99 & 3.61 & 3.80 \\
\texttt{+guided}	&	\emocell4.22 & 3.40 & \emocell4.57 & 3.89 & \emocell4.26 & 2.54 & 4.59 & 4.11 & \emocell4.41 & 3.49 & 3.95 \\
\texttt{+annotation}	&	4.14 & 3.52 & 4.54 & 3.60 & 4.23 & 2.66 & 4.48 & 3.99 & 4.35 & 3.44 & 3.89 \\
\texttt{+cga}	&	4.13 & \cultcell3.93 & 4.48 & \cultcell4.18 & 3.98 & 2.58 & \emocell4.64 & \cultcell4.38 & 4.31 & \cultcell3.77 & \textbf{4.04} \\
\bottomrule
\end{tabular}
}
\caption{Automatic evaluation results for all adaptation strategies and models used in our experiments. The ``All'' column is the average between emotional supportiveness and cultural awareness. Note that \texttt{redditor} here refers to the baseline strategy where the LLM plays the role of a Reddit responder, not the actual human-redditor response.} 
\label{tab:main}
\end{table*}

\begin{figure*}[h]
    \centering
    \includegraphics[width=0.84\linewidth]{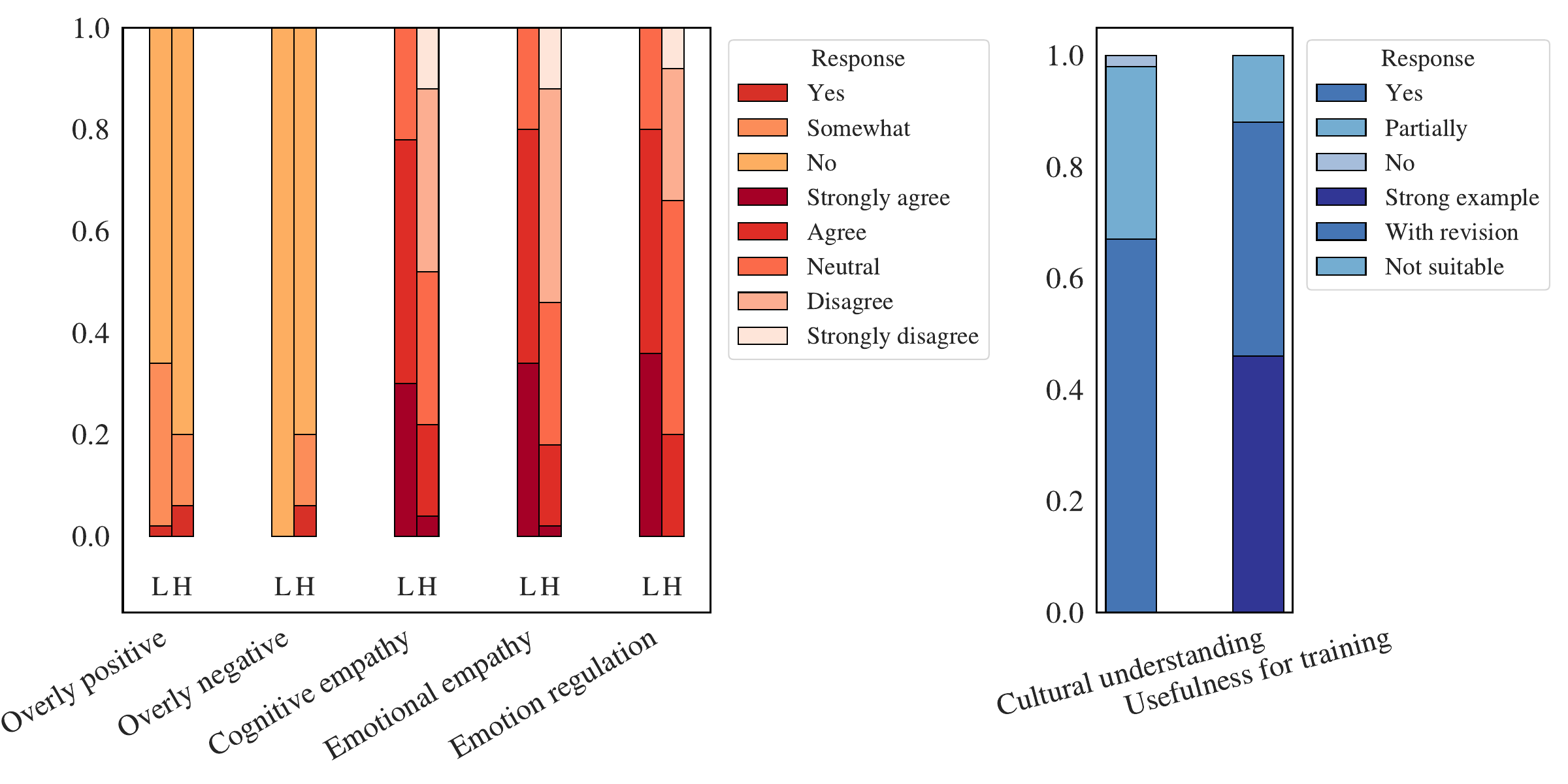}
    \caption{Adapted LLM responses are safe, culturally aware, and suitable for training clinical psychologists in cross-cultural therapy. \textbf{L}: LLM response, \textbf{H}: Human response.}
    \label{fig:expert_results}
\end{figure*}

\subsection{Automatic Evaluations}

We perform fine-grained automatic evaluations based on both emotional support quality and cultural awareness. We use GPT-o3-mini \cite{o3mini} as the LLM-as-a-Judge due to its high ability for reasoning and correlation with human judges \cite{iclr/TanZMTC0PS25/llmjudge, corr/abs-2411-15594/llmjudge}.

\noindent\textbf{Emotional supportiveness} measures the basic requirements of an effective supporting message. Based on evaluation criteria from prior research \cite{naacl/rashkin-etal-2019-towards/empatheticdialog, liu2021towards}, we included the following criteria: 
1. \emph{Empathy} - the response should demonstrate a genuine understanding of the author's emotions and convey timely, appropriate concern; and 2. \emph{Helpfulness} - the response offers effective advice and tailored, actionable steps. 

\noindent\textbf{Cultural awareness} measures the awareness and sensitivity of a response concerning cultural aspects. To match the desirable culturally sensitive support in \citet{cccir}, three criteria are used: 1. \emph{Socio-political influence} - the response demonstrates an understanding of the current sociopolitical system and its impact on the author of the post; 2. \emph{Knowledge} - the response reflects knowledge about the target culture; and  3. \emph{Cultural context} - the response perceives problem within the appropriate cultural context.

\noindent\textbf{Language quality} metrics are also derived from \citet{liu2021towards} and \citet{cccir}, measures: 1. \emph{Fluency} - the response should be coherent and easy to understand; 2. \emph{Communication} - the response is appropriate.

We evaluate all aspects on a 5-point scale, which is a common setup in prior work \cite{naacl/rashkin-etal-2019-towards/empatheticdialog, liu2021towards}. Following the evaluation prompt-generation method in G-Eval \cite{geval/emnlp/liu-etal-2023-g}, we use ChatGPT to create prompts that consist of step-by-step evaluation guidelines. See Appendix~\ref{app:eval_prompts} for all prompts. 

\subsection{Human Evaluations}
\label{sec:humaneval}
To further assess the emotional supportiveness and cultural awareness of the responses, we conducted two human evaluations: 1. \underline{Crowd} evaluations with individuals from the corresponding culture; 2. \underline{Expert} evaluations to assess the responses for safety and potential usefulness for professionals.

\noindent\textbf{Crowd evaluation.} To compare the effectiveness of different strategies and LLMs, we conducted in-culture crowd-sourced evaluations using Prolific.\footnote{\url{https://www.prolific.com/}} We recruit two people per culture who are fluent in both English and the matching language of the culture.\footnote{Most Chinese cultural data comes from individuals in English-speaking countries who identify with Chinese culture, shaping their distress experiences. We selected crowd evaluators to match this context.} 

\underline{\textit{Best adaptation strategy.}} To evaluate the best adaptation strategy, we sampled 30 posts and corresponding responses for every culture and LLM examined. This results in 2880 evaluations in total.
For each post, we display both the human and the model-generated responses. For each evaluation instance, we ask two evaluators to pick the best response in terms of both emotional supportiveness and cultural awareness, then aggregate their ranks of adaptation strategies.

\underline{\textit{Best LLM.}} Next, to evaluate the best model, we sampled 20 posts and corresponding responses from the best-performing strategy (\texttt{+cga}), and recruited 3 annotators per culture to find out the best-performing LLM. This results in 720 evaluations in total.
Here, we use the same criteria as the previous evaluation. The labels are decided based on a majority vote. If there is a tie, another annotator from the same culture is involved to make the judgment.\footnote{This only happens 5\% for cultural awareness, 13.33\% for emotional supportiveness, indicating that cultural awareness is a less subjective task than emotional supportiveness.} The instructions were given in both English and the native language of the culture. The detailed instructions are in the Appendix \ref{app:crowd_eval}.

\noindent
\textbf{Expert evaluation.} We collaborated with two psychologists experienced in online emotional support to validate the safety of the adaptation strategies (including cognitive and emotional empathy) and assess their utility for training psychology students in cross-cultural therapy (\citealt{benuto2018training}, including cultural understanding and usefulness for teaching cultural competence). Culturally sensitive emotional support is a well-established component of psychology education, and our use of LLMs reflects real needs identified by practitioners. Together, we developed evaluation guidelines (Appendix~\ref{app:expertguide}) and evaluated 200 responses, illustrating one way our work could support real-world training and educational applications.

\begin{table*}
\centering
\resizebox{0.8\linewidth}{!}{

\begin{tabular}{l cc cc cc cc}
\hline
\textbf{Model} & \multicolumn{2}{c}{\textbf{Arabic}} & \multicolumn{2}{c}{\textbf{Chinese}} & \multicolumn{2}{c}{\textbf{German}} & \multicolumn{2}{c}{\textbf{Jewish}}  \\
 & Emo. & Cult.  & Emo. & Cult.  & Emo. & Cult. & Emo. & Cult.  \\
\hline
Aya-Expanse-8B & \greencell\texttt{+cga} & \greencell\texttt{+cga} & \texttt{+a} & \texttt{+a} & \texttt{+a} & \texttt{+a} & \texttt{+a} & \texttt{+a} \\
Qwen-2.5-7B & \greencell\texttt{+g} & \greencell\texttt{+cga} & \texttt{+a} & \greencell\texttt{+cga} & \texttt{+cga} & \texttt{+cga} & \greencell\texttt{+a} & \greencell\texttt{+cga} \\
Llama-3.1-8B & \texttt{+cga} & \greencell\texttt{+cga} & \greencell\texttt{+g} & \greencell\texttt{+cga} & \greencell\texttt{+g} & \texttt{+a} & \greencell\texttt{+cga} & \greencell\texttt{+cga} \\
\hline
\end{tabular}
}
\caption{In-culture human evaluation results for the best strategies. \texttt{+a} is the \texttt{+annotation} strategy and \texttt{+g} is the \texttt{+guided} strategy. Overall, \texttt{+cga} and \texttt{+annotation} are the top-ranking strategies by humans. The shaded cells indicate human preferences match the best strategy by automatic evaluations from the model.}
\label{tab:human_eval}
\end{table*}

\section{Results \& Discussions}
\label{sec:results}

Table \ref{tab:main} shows the automatic evaluation results for both the emotion supportiveness and cultural awareness aspects. Across models, the strategy showing the best culturally-aware responses (i.e., blue boxes) is \texttt{+cga}, demonstrating the effectiveness of combining culture-informed role-playing, cross-cultural-competence guidelines and explicit cultural signal annotations. Moreover, these results also show that simple culture-informed role-playing alone (\texttt{+culture}) is not enough for offering the most culturally-aware responses.   In fact, \texttt{+culture} performs worse than providing explicit, detailed cultural signals (\texttt{+annotation}) or guidelines aimed at \emph{cross-cultural} consultation (\texttt{+guidelines}) for both emotional supportiveness and cultural awareness. By explicitly incorporating cultural considerations when offering emotional support (e.g., through \texttt{+guided}, \texttt{+annotation} or more significantly through  \texttt{+cga}), models generate better responses compared to \texttt{+culture} in both evaluation dimensions. Additionally, we observe similar trends reflected in the human evaluation results (Table~\ref{tab:human_eval}). More details on the human evaluation are in the next subsection and in \S\ref{subsec:human_eval_results}. We also observe consistent patterns on two additional 70B-parameter models (Llama and Qwen, Table \ref{tab:main_llm}; Appendix \ref{app:add_auto_eval}).

The automatic evaluation results for the language quality of the responses display nearly perfect scores across strategies and models (Table~\ref{tab:app_oq}), we therefore focus on emotional supportiveness and cultural awareness evaluation in the remainder of this work .

\subsection{Human (Crowd) Evaluation}\label{subsec:human_eval_results}

\noindent
\textbf{Adding cultural annotations is essential for more culturally-aware emotional support.} As shown in Table~\ref{tab:human_eval}, across all LLMs and all cultures, the winning strategy for the cultural aspect is a mix of \texttt{+a} and \texttt{+cga}, noting that for both strategies, the cultural signals are included in the input. This shows that high-quality culture annotations can surely enhance the cultural awareness of an emotional support message.

\noindent
\textbf{Human and LLM show moderate to strong correlations.} We compute the Kendall rank correlation coefficient ($\tau$) between humans' and LLMs' ranking of adaptation strategies. For Arabic and Chinese, human-model correlations, averaged over LLMs, are moderate to strong for emotional awareness, namely $\tau$=0.8 for Arabic and 0.66 for Chinese. For German and Qwen's responses to the Jewish culture, it is, however, low.  
Correlations for cultural awareness are weaker (e.g., 0.6 for Arabic and 0.47 for Chinese, averaged over LLMs).  However, as shown in Table \ref{tab:human_eval}, humans and the LLM-as-a-Judge often agree on the best strategy per model, suggesting that divergences mainly stem from lower-ranked strategies, i.e., humans and LLMs agree on the best but not the worst. The detailed results are in Table \ref{tab:correlations} (Appendix~\ref{app:human_eval}).

\subsection{Evaluation with Clinical Psychologists}
\noindent 
\textbf{LLMs do not escalate nor introduce new emotional distress.} Psychologists agree that LLMs' responses \emph{do not} introduce new distress or escalate existing negative feelings with aggregated positive rating (``Strongly agree'' and ``Agree'') 80\% of the time, compared to 28\% for Reddit responses.

\noindent 
\textbf{Adapted responses support training.} Psychologists found 88\% of LLM responses promising for cross-cultural therapy training, with 46\% rated as strong examples and 42\% requiring minor adjustments, showing their potential utility in educational contexts.

More details are in Figure~\ref{fig:expert_results} and Appendix~\ref{app:expert_results}.

\noindent 
\textbf{Qualitative insights.} LLMs generate structured, culturally sensitive, and empathetic responses, often including clear guidance, validation, and reassurance. Human replies, while more spontaneous and authentic, sometimes lack consistency or depth in addressing cultural and emotional needs. Experts noted ethical considerations, such as the need for transparency when users interact with AI, the risk of missing clinical red flags, and questions about whether models should closely mimic humans or retain a distinct AI voice. These findings suggest that, in controlled settings, LLMs can complement human support by providing reliable, culturally informed guidance while highlighting areas where human judgment remains essential.~\footnote{We emphasize that these results are specific to our dataset and evaluation context, which involved brief, reactive online peer responses, and do not imply a global superiority of LLMs over human supporters.}

\section{Conclusion and Future Work}
The ability of LLMs to provide culturally sensitive peer emotional support has been largely overlooked. To address this, we introduce \methodname{}, a fine-grained dataset spanning four cultures, and evaluate three state-of-the-art LLMs with multiple adaptation strategies. Our results show that shallow cultural cues are insufficient, while contextualized, guideline-aligned adaptations substantially improve performance. Collaborating with professional psychologists, we demonstrate the potential of culturally adapted LLMs for training psychology students. For this emerging research area, we focus on single-turn interactions to establish a reliable foundation for evaluation, with future work extending to multi-turn dialogues, broader cultural contexts, and real-world applications.

\section{Limitations}
\noindent
\textbf{Data source.} In this paper, we investigate culturally aware emotional support using data from Reddit. We acknowledge that the platform, its users and annotations may introduce representational biases, and thus do not offer a comprehensive representation of any particular culture. In future work, we aim to collect a more diverse and representative dataset through alternative sources and extensive large-scale annotations. 

\noindent
\textbf{Dataset size.} Our dataset contains 462 posts, which is smaller than many generic emotion support datasets. However, it includes over 4,000 fine-grained annotations created with in-culture annotators and quality checkers, making it the largest and most detailed resource of its kind. While its size may limit certain large-scale analyses, we believe the dataset's cultural depth and annotation quality make it highly valuable for evaluating LLMs. We discuss this in detail in Section~\ref{sec:462}.

\noindent
\textbf{Culture and language.} We also attempt to move away from the common practice of using language (which is often determined by data availability) or nationality (which is typically unavailable in anonymous online communities) as a boundary for cultures. While our approach has certain limitations, it also offers a novel contribution by focusing on self-identified cultural identity and on the underlying causes of emotional stress due to cultural factors as expressed by the users themselves.

\noindent
\textbf{Cultural coverage and model bias. } We acknowledge that models can exhibit cultural biases, and fully addressing these biases across all cultures remains an open challenge. In this work, we explored adaptation for four cultures (Jewish, Arabic, Chinese, and German), showing that culturally informed prompting and the incorporation of explicit cultural signals can improve culturally sensitive emotional-support responses in online settings, though these findings may generalize to other contexts only with careful attention and validation.

\section*{Ethics Statement}
All data used in this study are publicly available posts from Reddit, an anonymous forum, ensuring user identities are not disclosed. Our work only collects posts, and the demographic information is voluntarily provided in the original posts on Reddit.  While this work explores promising strategies for automatically providing culturally aware emotional support, we do not recommend using our method directly without human verification and large-scale robustness and safety testing. This research is approved by the ethics committee of the Technical University of Darmstadt (EK 121/2024).

 \section*{Acknowledgements}

 This work was supported by the DYNAMIC center, which is funded by the LOEWE program of the Hessian Ministry of Science and Arts (Grant Number: LOEWE/1/16/519/03/09.001(0009)/98). This work has also been funded by the LOEWE Distinguished Chair ``Ubiquitous Knowledge Processing'', LOEWE initiative, Hesse, Germany (Grant Number: LOEWE/4a//519/05/00.002(0002)/81).

 We thank Yael Bar-Shacha for her help and suggestions on our expert evaluation. We thank Thy Thy Tran, Doan Nam Long, Aishik Mandal, and Anmol Goel for their feedback on a draft of this paper.

\bibliography{custom}
\bibliographystyle{acl_natbib}

\clearpage
\appendix
\section{Additional Information about \methodname{}}
\label{app:dataset}

Figure \ref{fig:creationpipeline} illustrates our overall dataset creation pipeline.

\subsection{Data License}
Our experiment code and annotations are publicly available for research evaluations only. Due to Reddit's privacy and data policy, please contact the authors or refer to the README in our project GitHub for details on how to access the dataset texts.

\subsection{Lists of Subreddits and Keywords}
\label{datacollection}

\begin{enumerate}
    \item \textit{Search culture subreddits with mental health keywords:}
    \begin{itemize}
        \item Subreddits: \verb|r/arabs|, \verb|r/algeria|, \verb|r/bahrain|, \verb|r/egypt|, \verb|r/iraq|, \verb|r/jordan|, \verb|r/kuwait|, \verb|r/lebanon|, \verb|r/libya|, \verb|r/morocco|, \verb|r/oman|, \verb|r/palestine|, \verb|r/qatar|, \verb|r/saudiarabia|, \verb|r/somalia|, \verb|r/sudan|, \verb|r/syria|, \verb|r/tunisia|, \verb|r/uae|, \verb|r/yemen|, \verb|r/china|, \verb|r/shanghai|, \verb|r/beijing|, \verb|r/germany|, \verb|r/depression_de|, \verb|r/de|, \verb|r/german|, \verb|r/Israel|, \verb|r/hebrew|, \verb|r/HebrewIsaelis|, \verb|r/Judaism|, \verb|r/Jewish|
        \item Keywords: depression, depressed, anxiety, anxious, bipolar, autistic, sad, mental health, trauma, schizophrenia, schizophrenic, anger issues
    \end{itemize}
    \item \textit{Search mental health subreddits with culture keywords:}
    \begin{itemize}
        \item Subreddits: \verb|r/depression|, \verb|r/mmfb|, \verb|r/anxiety|, \verb|r/bipolar|, \verb|r/bpd|, \verb|r/autism|, \verb|r/schizophrenia|, \verb|r/mentalhealth|, \verb|r/traumatoolbox|, \verb|r/socialanxiety|, \verb|r/anger|, \verb|r/offmychest|, \verb|r/bodyacceptance|
        \item Keywords: arab, german, chinese, jewish + translated mental health keywords (e.g., traurig for sad in German)
    \end{itemize}
\end{enumerate}

The \texttt{restrict\_sr} parameter was set to false in the API configuration to prevent query filtering strictly by subreddit. This allows the retrieval of relevant posts from across Reddit \textit{when no matches are found within the targeted subreddit}, increasing data coverage and flexibility.

\subsection{Rule-based and LLM-based Filtering}
\label{filtering}

Our rule-based filters removed 46\% of the noisy instances. The filters are:
\begin{itemize}
    \item \textit{F1 - incomplete posts:} the post has a title but empty content.
    \item \textit{F2 - no personal distress message:} the post does \textit{not} contain a personal emotional distress message, i.e., there is no use of the words ``I'', ``me'', etc. Such posts, for example, discuss distress of other people or a tragic event in the news.
    \item \textit{F3 - incorrect language:} we restrict our data to English and the languages of our four cultures.
    \item \textit{F4 - more links and numbers than text:} we notice that posts with too many URLs and telephone numbers tend to provide information and pointers on seeking and undestanding mental health and \textit{does not express personal distress.}
\end{itemize}

Our LLM-based filters removed another 46\% of the data instances. We run the filters on Qwen-2.5-72B, The filters are:
\begin{itemize}
    \item \textit{F5 - no emotional distress detected:} an LLM-based classifier to rule out a post if it does not contain a personal distress message. (an advanced version of F2).
    \item \textit{F6 - wrong or no cultural context: } we instruct an LLM to rule out a post if it was not culturally relevant for our four cultures. 
\end{itemize}

\subsection{Annotation Guidelines}
\label{app:anno_guidelines}

We developed our annotation guidelines over iterations, and the final guidelines are in Figure \ref{fig:annotation_guidelines}.

\noindent\textbf{Posts.} For emotional intensity annotation, we use a 3-point scale for simplicity: 1-light, 2-moderate, 3-high. For detailed definitions and examples, see Table \ref{tab:intensity}.

For cultural signals annotation, we follow a modified version of the taxonomy of cultural elements based on \cite{tacl/abs-2406-03930/culture}, with 7 categories: Concepts, Knowledge, Values, Norms and Morals, Language, Artifacts, and Demographics. For detailed definitions of each category and examples, see Table~\ref{tab:cultureschema}.

\noindent\textbf{Responses.} For emotional responses, we annotated both support message strategies and empathy scores. 

Following ESConv \citep{liu2021towards}, the support message strategies span 8 categories: Questions, Restatement or Paraphrasing, Reflection of Feelings, Self-disclosure, Affirmations and Reassurance, Providing Suggestions, Information, and Others. For detailed definitions of each category and examples, see Table~\ref{tab:supportchema}.

We used a 5-point scale \citep{liu2021towards} to annotate the empathy scores of the responses. For detailed definitions and examples, see Table~\ref{tab:empathy}.

To protect annotators' mental health, we implemented the following measures: 1. Full transparency about the content, with clear warnings and labels for discussions involving mental distress, abuse, PTSD, and related topics. 2. Allowing annotators to terminate the task at any time.

\begin{figure*}[t]
\begin{tcolorbox}[title={\methodname{} Annotation guidelines}]
First, open your assigned annotation sheet. The first two rows of the annotation sheet are reserved for headings and instructions. Do not change them!\\

Every row contains one post-response pair. Finish one before moving to the next pair.\\

\begin{enumerate}
    \item Read the Reddit post written by OP (Author of the post).
    \item Identify and rate the intensity of personal emotional distress messages in the post. Details on intensity ratings are in Table~\ref{tab:intensity}.
    \item Identify and classify the cultural signals in the post. Cultural signals schema is in Table~\ref{tab:cultureschema}.
    \item Read the response.
    \item Identify and classify the emotional support messages in the response. The possible support strategies are listed in Table~\ref{tab:supportchema}.
    \item Identify and categorize the cultural signals in the response.
    \item Rate the empathy in the response. Details are in Table~\ref{tab:empathy}.
    
\end{enumerate}

\end{tcolorbox}   
\caption{Instructions for the annotation guidelines.}
\label{fig:annotation_guidelines}
\end{figure*}

\begin{table*}[t]
\centering
\begin{tabular}{l p{6cm} p{6cm}}
\toprule
  \textbf{Rating} & \textbf{Explanation} & \textbf{Example}\\
\hline
1 (light) & The emotion is present but subtle, with mild expression or little emphasis. & I had a bit of a rough day. \\ 
2 (moderate) & The emotion is clearly expressed, showing a noticeable impact without being overwhelming. & Today was pretty hard; I'm feeling down and could use some cheering up. \\ 
3 (high) & The emotion is intense and strongly emphasized, often reflecting deep or overwhelming feelings. & I'm absolutely devastated. I can't believe this happened, and I don't know how to cope.\\  
\bottomrule
\end{tabular}
\caption{Intensity scale of the post.}
\label{tab:intensity}
\end{table*}

\begin{table*}
\centering
\begin{tabular}{l p{6cm} p{6cm}}
\toprule
  \textbf{Category} & \textbf{Explanation} & \textbf{Example}\\
\hline
Concepts & Basic units of meaning underlying objects, ideas, or beliefs. & Familismo in Mexican culture. It reflects the belief that family comes first, and extended family members are often involved in decisions and provide emotional and practical support. \\ 
Knowledge & Information that can be acquired through education or practical experience. & Five pillars of Islam \\ 
Values & Beliefs, desirable end states or behaviours ranked by relative importance that can guide evaluations of things. &Respect for elders in Chinese culture\\  
Norms and Morals & Set of rules or principles that govern people's behaviour and everyday reasoning. & ``You should behave like a happy and thankful child''\\
Language & Specific use of slang, speech, dialects. & ``That sounds so sick!''\\
Artifacts & Materialized items as the productions of human culture, they can be forms of art, tools, machines, etc. & The Christian cross\\
Demographics & Talking about nationality or ethnicity. & ``We are a Chinese couple''\\
\bottomrule
\end{tabular}
\caption{Categories of cultural signals, adapted from the taxonomy in \citet{tacl/abs-2406-03930/culture}.}
\label{tab:cultureschema}
\end{table*}

\begin{table*}
\centering
\resizebox{0.99\linewidth}{!}{
\begin{tabular}{p{4cm} p{6cm} p{6cm}}
\toprule
  \textbf{Strategy} & \textbf{Explanation} & \textbf{Example}\\
\hline
Questions & Questions related to mental health status & How often do you feel this way? \\ 
Restatement or Paraphrasing & Describe the situation in other words. & Sounds like your family life is really tough. \\ 
Reflection of Feelings & Show empathy for the other person. Understand the other's feeling and behaviour. &I understand how anxious you are.\\  
Self-disclosure & Both have made the same experience and share the same feelings.&I feel the same way sometimes.\\
Affirmation and Reassurance & Help with the other person's uncertainty.&You have done your best and I believe you will get it.\\
Providing Suggestions & Give some possible solutions. &Find a responsible adult that you can confide in, someone you trust.\\
Information & Provide proved and useful information that can help to increase the mental well-being. &Apparently, lots of research has found that getting enough sleep before an exam can help students perform better.\\
Others & Remaining strategies. Includes wishes, hopes, etc. &I hope your luck will change.\\
\bottomrule
\end{tabular}
}
\caption{List of emotional support strategies, based on categories presented in ESConv \citep{liu2021towards}.}
\label{tab:supportchema}
\end{table*}

\begin{table*}
\centering
\resizebox{0.99\linewidth}{!}{
\begin{tabular}{l p{6cm} p{6cm}}
\toprule
  \textbf{Rating} & \textbf{Explanation} & \textbf{Example}\\
\hline
1 (not empathetic at all) & Shows no interest in others' feelings and may even dismiss or ignore them. & Everyone has problems. Just get over it. \\ 
2 (slightly empathetic) & Shows minimal acknowledgement of others' feelings, but lacks genuine concern or involvement. & Oh, that's unfortunate. \\ 
3 (moderately empathetic) & Recognizes others' emotions and shows some level of concern, but doesn't fully engage. & That sounds tough. I'm sorry you're dealing with that.\\  
4 (quite empathetic) & Actively listens, responds with understanding, and expresses care for the other person's emotions. & I can see why you feel that way. That must be really hard.\\  
5 (very empathetic) & Fully connects with and validates others' emotions, offering deep understanding and support. & I'm here for you, and I really understand what you're going through. Let me know how I can support you.\\  
\bottomrule
\end{tabular}
}
\caption{Empathy scores.}
\label{tab:empathy}
\end{table*}

\subsection{Language Statistics}
\label{app:language}

In this work, we deliberately chose not to use language as the defining boundary of culture, recognizing that culturally influenced distress can be expressed in any language in online communities. As a result, \methodname{} includes posts both in English and in the native languages associated with each culture. The language distribution varies by culture, and Table \ref{tab:language} shows the language statistics. 

\begin{table}
\centering
\resizebox{0.99\linewidth}{!}{
\begin{tabular}{lcc}
\toprule
  \textbf{Culture} & \textbf{English Only} & \textbf{Native Lang. / Code-mixing}\\
\hline
Arabic & 81 & 29 \\
Chinese & 131 & 10 \\
German & 19 & 100 \\
Jewish & 89 & 3 \\
\bottomrule
\end{tabular}
}
\caption{Language distributions in \methodname{}.}
\label{tab:language}
\end{table}

\subsection{Demographic Information}
\label{app:demographicinfo}

We extracted demographic information from posts using the prompt in Figure~\ref{fig:demographic_extractions} with GPT-4o-mini and manually validated the results. We follow a set of demographic attributes from the WVS \cite{haerpfer2022wvswave7}, which are included in Table \ref{tab:demo_category}. In some cases, only high-level cultural indicators, such as ``Arabic'' or ``Chinese'' are extractable.

\begin{table}[]
    \centering
    \begin{tabular}{l}
    \toprule
    \textbf{Attributes} \\
    \midrule
    Settlement \\
    Gender \\
    Age \\
    Born in \\
    Marital status \\
    Number of people in household \\
    Education \\
    Profession \\
    Employment \\
    Class \\
    Religion\\
    \bottomrule
    \end{tabular}
    \caption{Attributes of demographic information extracted in our work. For definition, see the prompt in Figure \ref{fig:demographic_extractions}.}
    \label{tab:demo_category}
\end{table}

\begin{figure*}[t]
    \centering
    \includegraphics[width=0.99\linewidth]{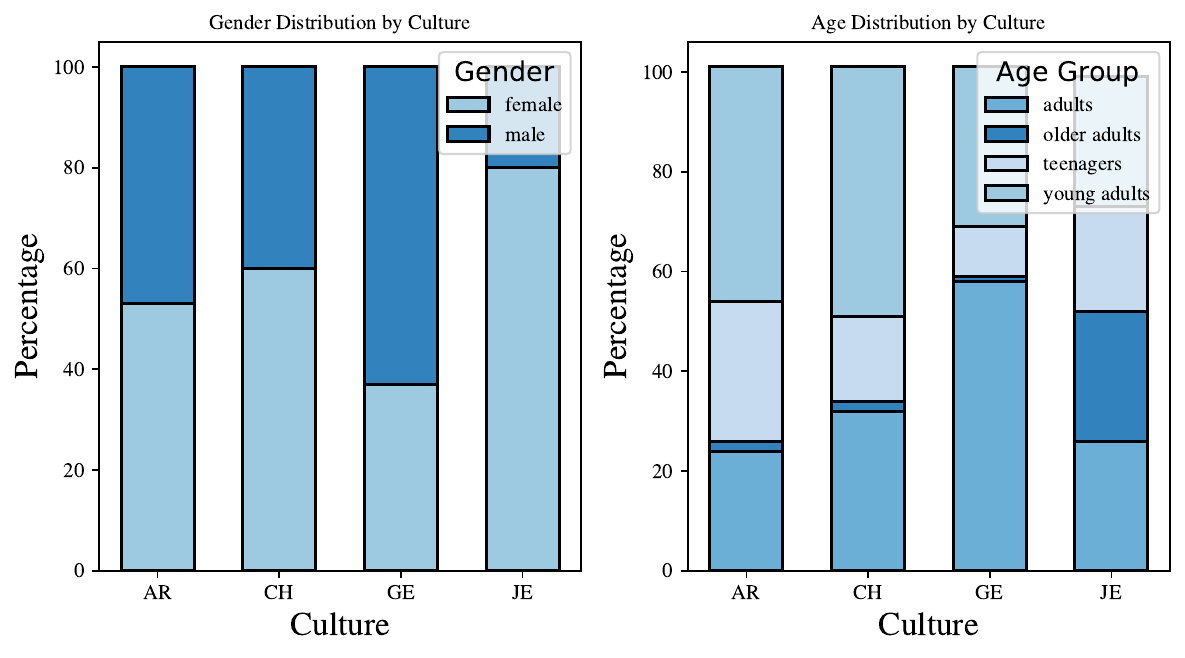}
    \caption{The gender and age group distributions across the four cultures of \methodname{}.}
    \label{fig:ageandgender}
\end{figure*}

Figure~\ref{fig:ageandgender} shows the distributions of age and gender in the \methodname{} dataset, computed over posts where this information is available (50\% of the dataset). The gender distribution varies notably across cultures: for instance, Jewish (JE) posts are predominantly female (80\%), while German (GE) posts are mostly male (63\%). Age distributions also differ; Arabic (AR) and Chinese (CH) have a higher proportion of young adults (47\% and 50\%, respectively), whereas GE and JE show a more balanced spread across adults and other age groups, with JE uniquely having an equal representation (26\%) in teenagers, young adults, adults, and older adults. These variations highlight cultural differences in demographic representation within the dataset. Table~\ref{tab:demoinfo} summarizes the demographic diversity in the \methodname{} dataset across the cultures. Notably, the dataset captures a wide variety of cultural backgrounds and life circumstances, from varied geographic origins and residences to diverse education levels and professions, reflecting rich cultural and social heterogeneity within each group.

\begin{figure*}
\begin{tcolorbox}[title={Demographic Information Extraction}]
You are an expert at structured data extraction. You will be given unstructured text from a Reddit post and should convert it into the given structure. Do not add any information that is not in the text. If there is no information, write 'unknown'. You should extract the following properties and return the output in JSON format:\\
\begin{verbatim}
{
  "settlement": "The location where the person lives.",
  "gender": "The gender of the person.",
  "age": "The age of the person.",
  "born_in": "The location where the person was born.",
  "marital_status": "The marital status of the person.",
  "people_in_household": "Household size.",
  "education": "The highest level of education of the person.",
  "profession": "The profession of the person.",
  "employment": "The employment status of the person.",
  "social_class": "The social class of the person.",
  "religion": "The religion or belief system of the person."
}
\end{verbatim}
  Here is the Reddit post: \{\texttt{post\_text}\} \\
    **Response**:
\end{tcolorbox}
\caption{Prompt for extracting demographic information from \methodname{}.}
\label{fig:demographic_extractions}
\end{figure*}

\subsection{Cultural Themes on Emotional Distress and Support}
\label{culturethemes}

Table~\ref{tab:normsandvalues} highlights how individuals from different cultural backgrounds experience emotional distress through the lens of cultural norms and values, the most prevalent cultural signals in our data (definitions are in Table~\ref{tab:cultureschema}). For example, both Arabic and Chinese distress messages report strong family control and dismissal of mental health issues, while Germans often express emotional exhaustion and social isolation. LGBTQ+ rejection appears uniquely in the Arabic context, while financial stress is more specific to the German entries. Gender expectations are prevalent in Arabic, Chinese, and Jewish contexts. Religious influence also varies significantly: in Arabic and Jewish cultures, religion plays a central role, either as a perceived cause of distress or as a coping mechanism. Overall, the table reveals that while emotional struggles are universal, the way they are shaped and expressed is deeply rooted in cultural values and social expectations.

\begin{table*}
\resizebox{0.99\linewidth}{!}{
\begin{tabular}{p{3.5cm} p{12cm}}
\toprule
\textbf{Culture} & \textbf{Attributes} \\
\hline
\textbf{Arabic} & gender: [male, female], age range: [14 to 40], birth place: [Iraq, arabic gulf countries, Arab country, Lebanon, Canada, Palestine, Middle east, U.S., Saudi Arabia, Syria], residence: [France, Sudan, California, UAE, Europe, Saudi Arabia, Dubai, Canada, West Bank, Iran, Aramoun (Lebanon), Bahrain, Syria, Nablus, Egypt], education: [college, high school, homeschooler, post graduate, middle school, bachelor's degree, master's degree], employment status: [employed, unemployed], marital status: [single, married], household size: [2-7 people], professions: [cybersecurity analyst, software engineer, pharmacist, medical worker, soldier, family business, ..], religion: [islam, christianity, atheism] \\
\hline
\textbf{Chinese} & gender: [male, female], age range: [15 to 58], birth place: [China, Taiwan, Malaysia, Shanghai, New Zealand, Cambodia, U.S., Europe, Australia], residence: [Fuzhou, Xingtai, Singapore, Shanghai, Malaysia, New Zealand, Hong Kong, Paris, China, Beijing, Texas, New York, Melbourne, Tokyo, North Carolina, U.K., Boston, North America], education: [high school, PhD, college graduate, master's degree, not completed, college junior], employment status: [employed, unemployed], marital status: [single, married, engaged, in a relationship, dating], household size: [1-4 people], professions: [graphic designer, BL artist, researcher, business expansion lead, English tutor, martial arts practitioner, florist, ..], religion: [christianity, buddhism] \\
\hline
\textbf{German} & gender: [male, female], age range: [16 to 42], birth place: [Germany], residence: [Germany, Berlin, Ostwestfalen, Horrem, Rheinland Pfalz, Dresden, Cologne, Kleinstadt], education: [Gesamtschule, German Abitur, vocational training, German Gymnasium, high school, Lehramtsstudium, Hauptschulabschluss, Bachelor], employment status: [employed, unemployed], marital status: [single, married, divorced, engaged], household size: [1-5 people], professions: [pizza delivery, social worker, caregiver, retail, optician, service worker, real estate agent, ..], religion: [rarely mentioned] \\
\hline
\textbf{Jewish} & gender: [male, female], age range: [12 to 47], birth place: [U.K., Former Soviet Union], residence: [Paris, Bay Area, New York, New Hampshire], education: [student, graduate school, college], employment status: [employed full-time, employed part-time], marital status: [married, single, engaged, separated, household size: [1-4 people], professions: [intern, project manager, artist, soldier, rabbi, nurse, ..], religion: [Judaism, Agnostic, Atheism, secular Jew, Ashkenazi Jew] \\
\bottomrule
\end{tabular}
}
\caption{Overview of the \methodname{}'s demographic diversity.}
\label{tab:demoinfo}
\end{table*}

\subsection{Types of Cultural Signals and Support Strategies per Culture}
\label{app:categoryperculture}

\begin{figure*}
    \centering
    \includegraphics[width=0.99\linewidth]{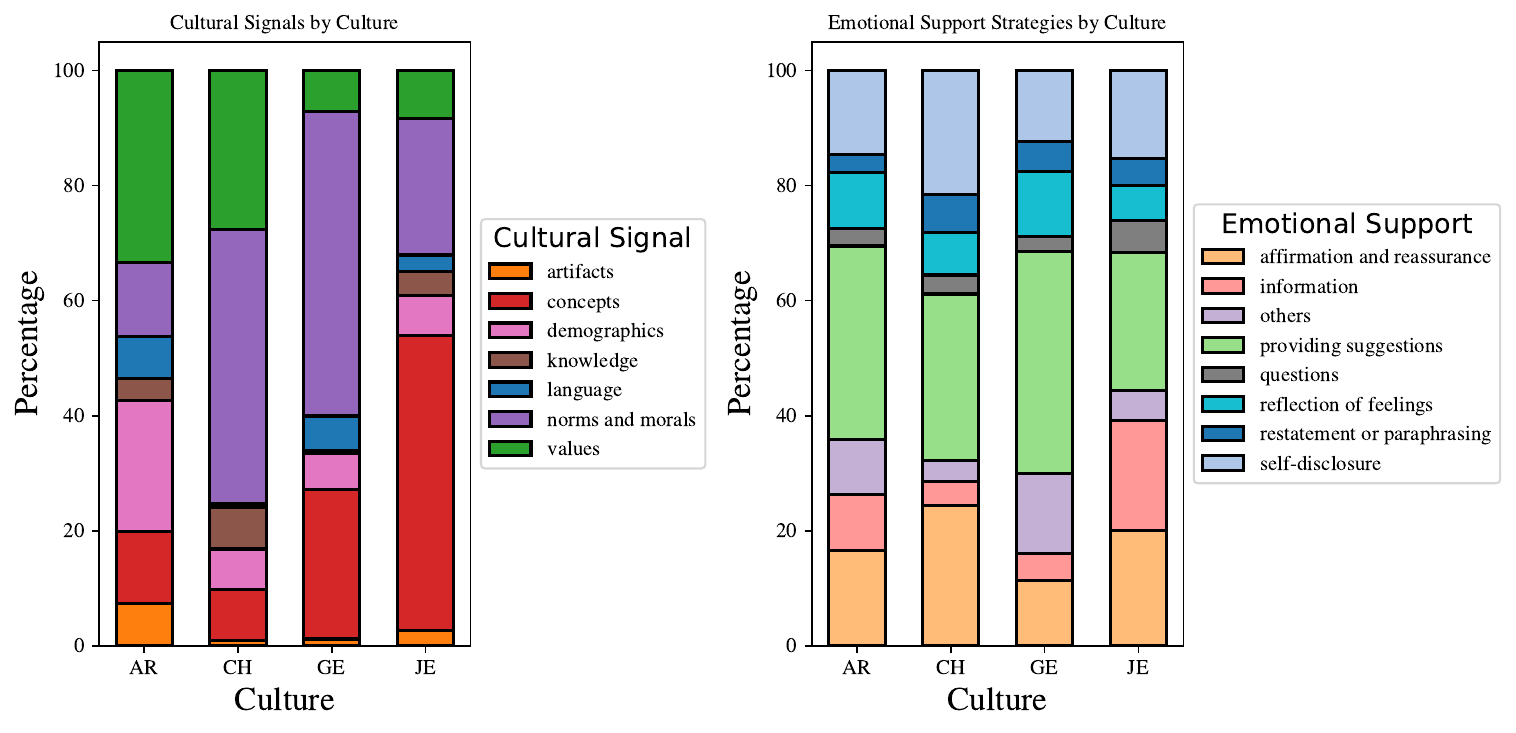}
    \caption{The types of cultural signals and support strategies for every culture in \methodname{}.}
    \label{fig:categoryperculture}
\end{figure*}

Figure~\ref{fig:categoryperculture} reveals the differences between the types of cultural signals among the 4 cultures. Each culture prioritizes certain signals over others, reflecting unique cultural characteristics. Arabic posts emphasize values and tend to state culture-specific demographics more explicitly (e.g., \textit{I'm an arab guy from Sudan}). Chinese and German posts prioritize norms and morals, indicating a strong focus on shared principles. In contrast, Jewish posts place a significant emphasis on cultural concepts (e.g., \textit{rabbi}, \textit{Shomer Shabbat}), with norms and morals playing a secondary but still important role. Across all cultures, signals of the types artifacts, knowledge, and language are minimally present.

The figure also illustrates cultural differences in emotional support strategies in human responses. The most prominent strategy in all cultures is providing suggestions. Beyond the top strategy, Arabic and Chinese, both emphasize self-disclosure and affirmation. German strategies are more balanced, and Jewish culture uses the information strategy more than the three other cultures. The least common strategy is asking questions. 

\subsection{Intensity of Emotional Distress and Empathy of Responses}
\label{app:empathyandintensity}
Table~\ref{tab:empathyandintensity} shows the average scores for the intensity of distress messages in posts and the empathy of human responses across cultures and overall. Jewish posts exhibit the highest intensity (2.07), followed closely by German posts (1.98), indicating that these cultures potentially express distress more vividly compared to Arabic and Chinese. In terms of empathy, Arabic responses show the highest level (3.27), followed by Jewish (3.18), suggesting a more explicit (verbal) approach to emotional support in these cultures. In contrast, Chinese responses demonstrate the lowest empathy (2.15), while German responses fall in the middle (2.73), reflecting a more moderate engagement. Overall, the data highlight cultural variability in both the expression of distress and empathetic responsiveness. 
It is important to note that since annotators evaluated posts and responses from within their own cultures, these scores may reflect culturally internalized norms. In future work, we plan to explore how perceptions shift when out-of-culture annotators assess the same content.

\begin{table}
\centering
\begin{tabular}{l l l l l l}
\toprule
  \textbf{Culture} & \textbf{Intensity} & \textbf{Empathy}  \\
\hline
\textbf{AR} & 1.81 & 3.27\\ 
\textbf{CH} & 1.77& 2.15\\ 
\textbf{GE} & 1.98 & 2.73\\ 
\textbf{JE} & 2.07 & 3.18\\ 
\textbf{All} & 1.89 & 2.77\\ 
\bottomrule
\end{tabular}

\caption{Average scores for intensity of distress messages in posts and empathy of human responses. \textbf{AR}: Arabic, \textbf{CH}: Chinese, \textbf{GE}: German, and \textbf{JE}: Jewish.}
\label{tab:empathyandintensity}
\end{table}


\subsection{Further Discussions on Dataset Size}
\label{sec:462}
While our dataset contains 462 posts, we argue that scale alone does not determine the usefulness of an evaluation benchmark. Prior work shows that LLM benchmarks with only a few hundred examples can still provide highly reliable insights. For example, HumanEval~\cite[164 problems]{chen2021evaluating} and DrawBench~\cite[200 instances]{saharia2022photorealistic} are widely used in state-of-the-art evaluations. Although smaller in absolute size than generic emotion support datasets, ours includes over 4,000 fine-grained annotations with in-culture annotators and quality checkers. As shown in Tables~\ref{tab:comparisons} and~\ref{tab:stats}, it is the largest of its kind and uniquely rich in cultural and emotional details. We further argue that an overemphasis on dataset size risks obscuring contributions that prioritize nuance, cultural specificity, and human well-being; context-rich resources such as ours provide insights that scale alone cannot, and are essential for comprehensive LLM evaluation.

\section{Infrastructure}
\label{app:infra}

We generate responses using the default temperature setting for all models with full precision. All our experiments were conducted using a single Nvidia A6000 GPU.

\section{Prompts}
\label{app:prompts}

\subsection{Adaptation Prompts}
The adaptation prompts for different strategies are outlined in Figures \ref{fig:culture_profile} to  \ref{fig:guided_annotationp2}. Prompts specify the response language. Even when simulating a cultural role, replies must match the post's language, e.g., Arabic if the post is in Arabic, English if in English by an Arab person.

\subsection{Auto-evaluations Prompts}\label{app:eval_prompts}
The evaluation prompts are outlined in Figures \ref{fig:eval_prompt} to  \ref{fig:eval_prompt8}. Our evaluation covers both emotional and cultural aspects, as well as the overall response quality. Metrics are in Table~\ref{tab:metric_def}.

\begin{figure*}
\begin{tcolorbox}[title={+culture}]
Respond to the text below as an English-speaking Redditor from \{\texttt{culture}\} culture. Only provide the response. Do not expose your role.  Do not add unnecessary marks like **Title** or **Response** in your output. Your response should be in the same language as the post.\\
Post: \{\texttt{post}\} \\
    **Response**:
\end{tcolorbox}
\caption{Prompt for the culture-informed role-playing strategy.}
\label{fig:culture_profile}
\end{figure*}

\begin{figure*} 
\centering
\begin{tcolorbox}[title={+guided}]
Respond to the text below as an English-speaking Redditor replying to a post.    \\
    Only provide the response. Do not expose your role. \\
    Do not add unnecessary marks like **Title** or **Response** in your output. \\
    Your response should be in the same language as the post.\\

    **Response Guidelines:**\\
    The advice you give should align with the following characteristics, please adhere to them throughout the conversation and refer back to them before sharing all of your responses:\\
        1. Value and respect cultural differences.\\
        2. Be comfortable with differences.\\
        3. Understand the current sociopolitical system and its impact on the author of the post.\\
        4. Demonstrate knowledge about the author of the post's culture. \\  
        5. Communicate appropriately to the author of the post.\\
        6. Perceive the problem within the appropriate cultural context of the author of the post.\\
        7. Acknowledge and be comfortable with cultural differences.\\
Post: \{\texttt{post}\} \\
    **Response**:
\end{tcolorbox}   
\caption{Prompt for the guided principles/constitutions strategy.}
\label{fig:guided}
\end{figure*}

\begin{figure*}[t] 
\centering
\begin{tcolorbox}[title={+annotation}]
Respond to the text below as an English-speaking Redditor replying to a post.\\
    Only provide the response. Do not expose your role. \\
    Do not add unnecessary marks like **Title** or **Response** in your output. \\
    Your response should be in the same language as the post.\\

    The following annotations for this post include phrases that highlight personal emotional distress and cultural signals.\\
    For each distress message, a rating is provided to indicate the intensity of the emotion expressed in the phrase. \\
    Additionally, each cultural phrase is classified as a specific type of cultural signal.\\
    When responding to the post, take the annotations into account to provide a reply that reflects empathy and cultural sensitivity.   \\ 

    **Definitions:**\\
    Personal Emotional Distress Messages:\\
        Psychological discomfort or suffering stemming from an individual's internal experiences, such as anxiety, sadness, or frustration.\\

    Emotion Intensity Ratings:\\
        Light: The emotion is present but subtle, with mild expression or little emphasis.\\
        Moderate: The emotion is clearly expressed, showing a noticeable impact without being overwhelming.\\
        High: The emotion is intense and strongly emphasized, often reflecting deep or overwhelming feelings.\\

    Cultural Signals:\\
        Behaviors, symbols, language, or practices that convey shared values, beliefs, or identities within a specific cultural group.\\
        Types of Cultural Signals:\\
            Concepts: Basic units of meaning underlying objects, ideas, or beliefs.\\
            Knowledge: Information acquired through education or practical experience.\\
            Values: Beliefs or desirable behaviors ranked by their relative importance, guiding evaluations and decisions.\\
            Norms and Morals: Rules or principles governing people's behavior and reasoning in everyday life.\\
            Language: Specific use of slang, speech, or dialects within the cultural context.\\
            Artifacts: Material items produced by human culture, such as art, tools, or machines.\\
            Demographics: References to nationality, ethnicity, or group identity.\\
Post: \{\texttt{post}\} \\
    Here are the annotations for this post: \\
    Personal distress phrase \{\texttt{i}\}: \{\texttt{phrase}\}\\
    Intensity of distress phrase \{\texttt{i}\}: \{\texttt{intensity}\}\\
    $...$\\ 
    Culture signal type \{\texttt{i}\}: \{\texttt{type}\}\\
    Culture phrase \{\texttt{i}\}: \{\texttt{phrase}\}\\
    $...$\\
    **Response**:
\end{tcolorbox}
\caption{Prompt for the explicit cultural signals strategy.}
\label{fig:guided}
\end{figure*}

\begin{figure*}[t] 
\centering
\begin{tcolorbox}[title={+cga (part one)}]
Respond to the text below as an English-speaking Redditor from \{\texttt{culture}\} culture replying to a post. \\
    Only provide the response. Do not expose your role. \\
    Do not add unnecessary marks like **Title** or **Response** in your output. \\
    Your response should be in the same language as the post.\\

    The following annotations for this post include phrases that highlight personal emotional distress and cultural signals.\\
    For each distress message, a rating is provided to indicate the intensity of the emotion expressed in the phrase. \\
    Additionally, each cultural phrase is classified as a specific type of cultural signal.\\
    When responding to the post, take the annotations into account to provide a reply that reflects empathy and cultural sensitivity.   \\ 

    **Definitions:**\\
    Personal Emotional Distress Messages:\\
        Psychological discomfort or suffering stemming from an individual's internal experiences, such as anxiety, sadness, or frustration.\\

    Emotion Intensity Ratings:\\
        Light: The emotion is present but subtle, with mild expression or little emphasis.\\
        Moderate: The emotion is clearly expressed, showing a noticeable impact without being overwhelming.\\
        High: The emotion is intense and strongly emphasized, often reflecting deep or overwhelming feelings.\\

    Cultural Signals:\\
        Behaviors, symbols, language, or practices that convey shared values, beliefs, or identities within a specific cultural group.\\
        Types of Cultural Signals:\\
            Concepts: Basic units of meaning underlying objects, ideas, or beliefs.\\
            Knowledge: Information acquired through education or practical experience.\\
            Values: Beliefs or desirable behaviors ranked by their relative importance, guiding evaluations and decisions.\\
            Norms and Morals: Rules or principles governing people's behavior and reasoning in everyday life.\\
            Language: Specific use of slang, speech, or dialects within the cultural context.\\
            Artifacts: Material items produced by human culture, such as art, tools, or machines.\\
            Demographics: References to nationality, ethnicity, or group identity.\\
    
    **Response Guidelines:**\\
    The advice you give should align with the following characteristics, please adhere to them throughout the conversation and refer back to them before sharing all of your responses:\\
        1. Understand the current sociopolitical system and its impact on the author of the post.\\
        2. Demonstrate knowledge about the author of the post's culture.   \\
        3. Communicate appropriately to the author of the post.\\
        4. Perceive the problem within the appropriate cultural context of the author of the post.\\

\end{tcolorbox}
\caption{Prompt for the combined strategy, part one. The prompt continues in Figure \ref{fig:guided_annotationp2}.}
\label{fig:guided_annotationp1}
\end{figure*}

\begin{figure*}[t] 
\centering
\begin{tcolorbox}[title={+cga (part two)}]
Post: \{\texttt{post}\} \\
    Here are the annotations for this post: \\
    Personal distress phrase \{\texttt{i}\}: \{\texttt{phrase}\}\\
    Intensity of distress phrase \{\texttt{i}\}: \{\texttt{intensity}\}\\
    $...$\\ 
    Culture signal type \{\texttt{i}\}: \{\texttt{type}\}\\
    Culture phrase \{\texttt{i}\}: \{\texttt{phrase}\}\\
    $...$\\
    **Response**:
\end{tcolorbox}
\caption{Prompt for the combined strategy, part two. This is the continuation of the prompt in Figure \ref{fig:guided_annotationp1}.}
\label{fig:guided_annotationp2}
\end{figure*}

\begin{figure*} 
\centering
\begin{tcolorbox}[title={Base Prompt for Auto-Evaluation}]
        You will be given a post and a response. \\
        Your task is to rate the response on the given metric. \\
        Make sure you read and understand these instructions carefully. Please keep this document open while reviewing, and refer to it as needed. \\
        **IMPORTANT** Follow the instructions and provide ONLY your score for the metric. Do not respond in a whole sentence and don't explain. \\

        Evaluation Criteria: \\
        \{\texttt{metric\_name}\} (1-5) - \{\texttt{metric\_definition}\} \\

        Evaluation Steps: \\
        \{\texttt{evaluation\_steps}\} \\

        Instruction: Respond with exactly one single digit (1-5) and nothing else. \\
        Constraints: No words, no explanations, no punctuation, just one digit. \\
        Final Reminder: If you do not comply, your response is invalid. \\

        Post: \\
         \{\texttt{post}\} \\

        Response: \\
        \{\texttt{response}\} \\

        Evaluation score: \\
\end{tcolorbox}   
\caption{Base auto-evaluation prompt. Metric names and definitions are in Table \ref{tab:metric_def}, and evaluation steps prompts are in Figure \ref{fig:eval_prompt2} - Figure \ref{fig:eval_prompt8}.}
\label{fig:eval_prompt}
\end{figure*}

\begin{table*}[]
    \centering
\resizebox{0.99\linewidth}{!}{
    \begin{tabular}{p{4cm} p{12cm}}
    \toprule
    \textbf{Metric} & \textbf{Definition} \\
        \midrule
        \emocell Empathy &  Measure the frequency and depth of empathy exhibited by the response.  Evaluate whether the response shows a genuine understanding of the post's emotions and whether its responses reflect timely and appropriate concern.\\
       \emocell Helpfulness & Evaluate the ability of the response to provide practical solutions and assistance during the dialogue. Consider whether the model offers effective advice and actionable steps tailored to the post's specific problems, such as emotional distress or requests for help. \\
       \midrule
       \cultcell Socio-political influence & The responder understands the current sociopolitical system and its impact on the author of the post.\\
       \cultcell Knowledge & The responder demonstrates knowledge about the author of the post's culture.\\
       \cultcell Cultural context & The responder perceives the problem within the appropriate cultural context of the author of the post.\\
       \midrule
        Fluency &  Is the response fluent and understandable? \\
        Communication & The responder's communication is appropriate for the author of the post.\\
    \bottomrule
    \end{tabular}
}
    \caption{Criteria used with the basic auto-evaluation prompt in Figure \ref{fig:eval_prompt}. Top: Emotional supportiveness. Middle: Cultural awareness. Bottom: Overall quality.}
    \label{tab:metric_def}
\end{table*}

\begin{figure*} 
\centering
\begin{tcolorbox}[title={Auto-Evaluation Steps Prompt - Empathy}]
        1. Read the Post and Response Carefully \\
            Begin by reading the post thoroughly to understand its emotional tone, context, and any expressed or implied concerns.\\
            Then, read the response to assess how well it addresses the emotions and concerns presented in the post.\\
        2. Identify Empathetic Elements in the Response\\
            Look for signs of empathy in the response, including:\\
                Acknowledgement: Does the response recognize and validate the emotions expressed in the post?\\
                Understanding: Does the response demonstrate a clear and accurate understanding of the individual's feelings and situation?\\
                Supportiveness: Does the response offer appropriate reassurance, concern, or support without being dismissive or overly generic?\\
        3. Assess the Depth of Empathy\\
            Evaluate how deeply the response connects to the emotions and context of the post:
            Does it feel genuine and considerate, or does it come across as superficial or robotic?\\
            Is the response tailored to the individual's situation, or is it overly broad and impersonal?\\
        4. Rate the Response on a Scale of 1-5\\
            Use the following scale to assign a score:\\
                1: The response shows little or no empathy. It fails to acknowledge emotions or provide any support.\\
                2: The response shows limited empathy. It may vaguely acknowledge emotions but lacks depth or sincerity.\\
                3: The response demonstrates moderate empathy. It recognizes emotions and offers some support, but it could be more thoughtful or personalized.\\
                4: The response is empathetic and considerate, addressing emotions effectively with only minor areas for improvement.\\
                5: The response is highly empathetic, deeply understanding and addressing emotions with genuine care and tailored support.\\
        5. Document Your Rating\\
            Only record your score and do NOT provide a whole sentence.\\
\end{tcolorbox}   
\caption{Auto-evaluation steps prompt - Empathy.}
\label{fig:eval_prompt2}
\end{figure*}

\begin{figure*} 
\centering
\begin{tcolorbox}[title={Auto-Evaluation Steps Prompt - Helpfulness}]
        1. Read the Post and Response Carefully\\
            Begin by reading the individual's post to fully understand their specific problems, emotional state, or requests for help.\\
            Read the response to evaluate how well it addresses the individual's concerns and provides solutions.\\
        2. Analyze the Practicality of the Response\\
            Examine whether the response offers effective and actionable solutions:
                Relevance: Does the response address the main concerns or requests expressed in the post?\\
                Actionable Steps: Are the suggestions or advice practical, clear, and feasible for the individual to implement?\\
                Specificity: Does the response avoid vague or generic advice by offering detailed and relevant steps?\\
        3. Assess the Assistance Provided\\
            Consider the depth of support offered in the response:\\
                Problem-Solving: Does the response provide a tangible path toward resolving the issues raised?\\
                Emotional Support: If the individual is in distress, does the response combine practical advice with empathetic and supportive language?\\
                Adaptability: Does the response show an understanding of the individual's unique situation and offer advice tailored to their needs?\\
        4. Rate the Response on a Scale of 1-5\\
            Use the following scale to assign a score:\\
                1: The response is unhelpful, providing no meaningful advice or assistance related to the individual's problem.\\
                2: The response offers limited or generic advice with minimal practical application to the specific issue.\\
                3: The response provides moderately helpful advice, but it may lack depth, specificity, or alignment with the individual's unique circumstances.\\
                4: The response is helpful, offering practical, relevant, and mostly actionable advice with minor room for improvement.\\
                5: The response is highly helpful, delivering clear, tailored, and actionable solutions that directly address the individual's concerns with exceptional clarity and support.\\
        5. Document Your Rating\\
            Only record your score and do NOT provide a whole sentence. \\
\end{tcolorbox}   
\caption{Auto-evaluation steps prompt - Helpfulness.}
\label{fig:eval_prompt3}
\end{figure*}

\begin{figure*} 
\centering
\begin{tcolorbox}[title={Auto-Evaluation Steps Prompt - Socio-political influence}]
        1. Read the Original Post Carefully\\
            - Identify any mention of the sociopolitical system or structures (e.g., racism, immigration policies, gender inequality, class issues, systemic barriers, etc.).\\
            - Understand how the author is affected by these systems — are they marginalized, angry, disillusioned, or seeking validation?\\
        2. Read the Response Carefully\\
            - Look for evidence that the responder acknowledges or engages with the sociopolitical context brought up in the post.\\
            - Determine whether the response aligns with or ignores the structural or systemic issues mentioned by the author.\\
        3. Assign a Socio-political influence Score (1-5)\\
            - Use the rubric below to rate the sociopolitical influence:\\
            Score	Description\\
                5	The response demonstrates deep understanding of the sociopolitical issues affecting the poster. It explicitly recognizes systemic factors and validates the poster's experience in that context.\\
                4	The response shows solid awareness of relevant sociopolitical dynamics but may not fully explore them. It still affirms the poster's struggle in a systemically grounded way.\\
                3	The response is neutral or superficial about sociopolitical context. It might acknowledge the poster's emotions but fails to meaningfully engage with systemic issues.\\
                2	The response minimizes or misses the sociopolitical context, offering platitudes or individualistic framing where structural understanding is needed.\\
                1	The response is ignorant, dismissive, or contradicts the sociopolitical reality expressed in the post. It may invalidate or erase structural struggles.\\
\end{tcolorbox}   
\caption{Auto-evaluation steps prompt - Socio-political Influence.}
\label{fig:eval_prompt4}
\end{figure*}

\begin{figure*} 
\centering
\begin{tcolorbox}[title={Auto-Evaluation Steps Prompt - Knowledge}]
        1. Read the Original Post Carefully\\
            - Identify explicit or implicit cultural references (e.g., language, religion, traditions, holidays, values, family structure, gender norms, immigration experiences, etc.).\\
            - Consider how the author's cultural identity shapes their experience or distress.\\
        2. Read the Response Carefully\\
            - Look for signs that the speaker understands, respects, or accurately refers to the author's culture.\\
            - Evaluate the specificity and accuracy of any cultural references or framing.\\
            - Check for stereotyping, assumptions, or inappropriate generalizations.\\
        3. Assign a Knowledge Score (1-5)\\
            - Use this rubric:\\
            Score	Description\\
                5	The response shows strong, accurate, and nuanced knowledge of the author's culture. It reflects deep familiarity and avoids stereotypes.\\
                4	The response demonstrates clear and respectful understanding of relevant cultural context, with some specific references or insights.\\
                3	The response shows general cultural sensitivity, but with limited or vague cultural specificity. No harmful assumptions, but also no strong insight.
                2	The response lacks cultural understanding, makes generic or shallow statements, or leans on simplified views of culture.\\
                1	The response includes inaccurate, stereotypical, or offensive assumptions about the author's culture, or ignores cultural relevance entirely.\\
\end{tcolorbox}   
\caption{Auto-evaluation steps prompt - Knowledge.}
\label{fig:eval_prompt5}
\end{figure*}

\begin{figure*} 
\centering
\begin{tcolorbox}[title={Auto-Evaluation Steps Prompt - Cultural context}]
       1. Read the Original Post Carefully\\
            - Identify cultural references, values, or assumptions implied or stated by the author (e.g., family honor, expectations, shame, religious pressure, collectivism vs. individualism).\\
            - Ask: Is the author's distress connected to their culture, community, or social roles?\\
            - Consider how the author's cultural context frames their problem (e.g., arranged marriage, filial duty, stigma of mental illness).\\
        2. Read the Response Carefully\\
            - Evaluate whether the speaker recognizes the cultural context and responds with that understanding.\\
            - Look for evidence of cultural framing, sensitivity to norms, and avoidance of ethnocentric judgments.\\
            - Watch out for universalizing, individualistic reframing, or dismissal of culturally specific experiences.\\
        3. Assign a Cultural Context Score (1-5)\\
            - Use this rubric:\\
            Score	Description\\
                5	The response clearly understands and respects the cultural context of the issue. The response is deeply grounded in the author's cultural framework.\\
                4	The response shows good awareness of cultural context, with minor gaps or generalizations, but overall respectful and relevant.\\
                3	The response offers generic or culturally neutral support. The response does not harm, but also doesn't fully engage with cultural context.\\
                2	The response minimizes, misinterprets, or overlooks cultural factors. Some framing may be inappropriate or culturally mismatched.\\
                1	The response dismisses, erases, or contradicts the cultural framework of the author's issue. Response may seem judgmental, ethnocentric, or harmful.\\
\end{tcolorbox}   
\caption{Auto-evaluation steps prompt - Cultural context.}
\label{fig:eval_prompt6}
\end{figure*}

\begin{figure*}[t]
\centering
\begin{tcolorbox}[title={Auto-Evaluation Steps Prompt - Fluency}]
        1. Read the Post and Response \\
            Carefully read the post provided by the individual to understand the context, tone, and intent behind the message. Then, read the potential response to evaluate how well it meets the criteria. \\
        2. Focus on Fluency \\
            Evaluate the response solely based on fluency, which means assessing the following: \\
                Grammar and Syntax: Is the response free from grammatical errors or awkward phrasing? \\
                Clarity: Is the message easy to read and understand? \\
                Naturalness: Does the response sound like it could naturally come from a human? \\
                Flow: Do the sentences connect smoothly without abrupt or disjointed ideas? \\
        3. Ignore Other Factors \\
            While evaluating, ignore elements like relevance, emotional support, or appropriateness. Only focus on the fluency of the response, not how well it aligns with the original post or its context. \\
        4. Rate the Response on a Scale of 1-5 \\
            Use the following scale to assign a score: \\
                1: The response is not fluent and difficult to understand (e.g., contains significant grammar issues or incomprehensible phrasing). \\
                2: The response has noticeable issues with fluency, but the meaning can still be understood with effort. \\
                3: The response is somewhat fluent but has minor awkward phrasing or grammar issues that may disrupt the flow. \\
                4: The response is fluent and mostly natural, with very minor issues that do not hinder understanding. \\
                5: The response is highly fluent, natural, and flows smoothly without any noticeable errors or awkwardness. \\
        5. Document Your Rating \\
            Only record your score and do NOT provide a whole sentence. \\
\end{tcolorbox}   
\caption{Auto-evaluation steps prompt - Fluency.}
\label{fig:eval_prompt7}
\end{figure*}

\begin{figure*} 
\centering
\begin{tcolorbox}[title={Auto-Evaluation Steps Prompt - Communication}]
       1. Read the Original Post Carefully\\
            - Note the emotional tone, urgency, and vulnerability expressed by the author.\\
            - Consider the author's demographic or cultural background if relevant, including what kind of communication style might be most appropriate (e.g., formal/informal, emotionally validating, calm and grounding, etc.).\\
            - Ask: What kind of support does this person seem to need right now?\\
        2. Read the Response Carefully\\
            - Assess whether the speaker's tone, language, and framing match the needs of the author.\\
            - Check for empathy, respect, and sensitivity.\\
            - Consider cultural appropriateness (e.g., directness, honorifics, collectivist vs. individualist framing).\\
            - Watch out for patronizing, detached, clinical, or inappropriate tones.\\
        3. Assign a Communication Score (1-5)\\
            - Use the rubric below:\\
            Score	Description\\
                5	The response is highly attuned to the author's emotional and cultural needs. Tone is supportive, appropriate, and sensitive, matching the situation.\\
                4	The response is mostly appropriate, with minor mismatches in tone or framing that don't seriously undermine support.\\
                3	The response is neutral or somewhat mismatched in tone. Shows effort to connect, but might feel off, generic, or not emotionally in sync.\\
                2	The response has a clear mismatch in tone or formality. May come across as unhelpful, awkward, or emotionally disconnected.\\
                1	The response is inappropriate, dismissive, or insensitive. Tone may be offensive, overly clinical, judgmental, or otherwise harmful.\\
\end{tcolorbox}   
\caption{Auto-evaluation steps prompt - Communication.}
\label{fig:eval_prompt8}
\end{figure*}

\section{Additional Automatic Evaluation Results}\label{app:add_auto_eval}

Table \ref{tab:app_oq} shows the automatic evaluation for overall quality.  Models consistently score high in all strategies. This is expected as modern LLMs exhibit language influences, especially in English and high-resource languages.

\begin{table}[]
    \centering
\resizebox{0.99\linewidth}{!}{
    \begin{tabular}{l cccc}
    \toprule
    \textbf{Model} & \textbf{AR} & \textbf{CH} & \textbf{GE} & \textbf{JE} \\
    \midrule
Aya-Expanse-8B 	&	4.99 & 4.99 & 4.97 & 4.98 \\	
\texttt{+culture}	&	4.94 & 4.99 & 5.00 & 4.99 \\	
\texttt{+guided}	&	5.00 & 5.00 & 4.99 & 4.99 \\	
\texttt{+annotation}	&	5.00 & 5.00 & 4.95 & 4.99 \\	
\texttt{+cga}	&	5.00 & 5.00 & 4.99 & 5.00 \\	
			\midrule
Qwen-2.5-7B	&	4.76 & 4.98 & 4.55 & 4.94 \\	
\texttt{+culture}	&	4.83 & 4.97 & 4.74 & 4.98 \\	
\texttt{+guided}	&	4.99 & 4.99 & 4.97 & 4.97 \\	
\texttt{+annotation}	&	5.00 & 5.00 & 4.92 & 4.98 \\	
\texttt{+cga}	&	4.77 & 4.99 & 4.40 & 4.99 \\	
			\midrule			
Llama-3.1-8B	&	4.39 & 4.95 & 4.79 & 4.89 \\	
\texttt{+culture}	&	4.30 & 4.93 & 4.63 & 4.95 \\	
\texttt{+guided}	&	4.63 & 4.96 & 4.83 & 4.97 \\	
\texttt{+annotation}	&	4.44 & 4.99 & 4.91 & 4.97 \\	
\texttt{+cga}	&	4.62 & 5.00 & 4.84 & 4.98 \\
\bottomrule
    \end{tabular}
    }
    \caption{Automatic evaluation results for the overall quality (culture, emotion, and language quality metrics scores). \textbf{AR}: Arabic, \textbf{CH}: Chinese, \textbf{GE}: German, and \textbf{JE}: Jewish.}
    \label{tab:app_oq}
\end{table}
In this paper, we mainly focused on examining models in the 7-8B parameter range, given their popularity, performance, and suitability for agentic systems \cite{arxiv/slm/abs-2506-02153}, while still remaining competitive. Here, we provide additional results for two larger models---Llama-3.1-70B and Qwen-2.5-72B---in Table \ref{tab:main_llm}.
Here, we observe similar patterns as with small models. Overall, the \texttt{+guided} strategy is more effective at providing winning emotional support, while \texttt{+cga} is better attuned to cultural sensitivity. The overall winning strategy with Qwen-2.5-72B is \texttt{+cga}, whereas for Llama-3.1-70B it is \texttt{+guided}. Simple culture-informed role-playing (\texttt{+culture}) does not substantially improve cultural awareness compared to \texttt{+cga}.
Interestingly, the best overall average automatic evaluation scores for both Llama-3.1-70B (4.05) and Qwen-2.5-72B (4.12) are lower than that of Aya-Expanse-8B (4.51). This indicates that a larger model may not perform better for our specific task, necessitating further research.
\begin{table*}[th]
\centering
\resizebox{0.9\linewidth}{!}{
\begin{tabular}{l ccc ccc ccc ccc ccc}
\hline
\textbf{Model} & \multicolumn{2}{c}{\textbf{Arabic}} & \multicolumn{2}{c}{\textbf{Chinese}} & \multicolumn{2}{c}{\textbf{German}} & \multicolumn{2}{c}{\textbf{Jewish}} & \multicolumn{3}{c}{\textbf{Average}} \\
 & Emo. & Cult.  & Emo. & Cult.  & Emo. & Cult. & Emo. & Cult.  & Emo. & Cult. & All \\
\hline
Llama-3.1-70B	&  & & &  &  &  && & &  & \\

\texttt{redditor}	&  3.82 & 3.47 & 4.07 & 3.53 & 3.86 & 3.11 & 4.39 & 4.07 & 4.03 & 3.54 & 3.79 \\
\texttt{+culture}	& 3.75 & 3.97 & 4.01 & 4.13 & 3.94 & 3.14 & 4.39 & 4.18 & 4.02 & 3.86 & 3.94 \\
\texttt{+guided}	& 4.25 & 3.99 & 4.24 & 4.01 & 4.05 & 3.13 & 4.47 & 4.27 & \textbf{4.25} & 3.85 & \textbf{4.05} \\
\texttt{+annotation}	& 4.08 & 3.92 & 4.05 & 3.94 & 3.78 & 3.12 & 4.48 & 4.14 & 4.10 & 3.78 & 3.94 \\
\texttt{+cga}	& 3.74 & 4.23 & 4.24 & 4.32 & 3.69 & 3.20 & 4.39 & 4.35 & 4.01 & \textbf{4.02} & 4.02 \\

\hline
Qwen-2.5-72B&  & & &  &  &  && & &  & \\

\texttt{redditor} & 4.27 & 3.30 & 4.22 & 3.23 & 4.14 & 3.04 & 4.33 & 3.86 & 4.24 & 3.36 & 3.80 \\
\texttt{+culture}	& 4.05 & 3.58 & 4.20 & 3.73 & 4.33 & 3.21 & 4.64 & 4.02 & 4.31 & 3.63 & 3.97 \\
\texttt{+guided}	& 4.49 & 3.58 & 4.60 & 3.58 & 4.71 & 3.13 & 4.76 & 3.82 & \textbf{4.64} & 3.53 & 4.08 \\
\texttt{+annotation}	& 4.18 & 3.38 & 4.34 & 3.47 & 4.17 & 3.02 & 4.48 & 3.80 & 4.29 & 3.41 & 3.85 \\
\texttt{+cga}	& 4.37 & 3.93 & 4.28 & 3.93 & 4.37 & 3.33 & 4.62 & 4.14 & 4.41 & \textbf{3.83} & \textbf{4.12} \\

\bottomrule
\end{tabular}
}
\caption{Automatic evaluation results for all adaptation strategies using 70B-parameter models. The ``All'' column is the average between emotional supportiveness and cultural awareness.  }
\label{tab:main_llm}
\end{table*}

\section{Crowd Evaluations}\label{app:crowd_eval}

\subsection{Instructions}
We show our crowd evaluation instructions in Figure \ref{fig:human_eval_guidelines1}. Both best adaptation strategy evaluation and best model evaluation use the same instructions, and we also translate the instructions into a language aligning with the culture. 

\begin{figure*}
\begin{tcolorbox}[title={Crowd Evaluation Instructions}]
Your task is to select the best response based on the following criteria separately:\\
	1. Emotional supportiveness\\
	2. Cultural awareness\\

\rule{\linewidth}{0.4pt}

What is an ``emotional supportive message''?	\\
Communication aimed at providing comfort, reassurance, and understanding to someone experiencing stress or emotional difficulty. The message might also include helpful suggestions.\\

What is a good emotional supportive message?	\\
1. the response should demonstrate a genuine understanding of the author's emotions and convey timely, appropriate concern. \\	
2. the response offers effective advice and tailored, actionable steps. \\

Good example phrases from responses for each criteria:\\	
1. Losing a loved one is never easy, and it's even harder when there's a language barrier between you and someone who needs your support.\\
2. One thing that might be helpful is to ask your sister or other family members if they can help facilitate communication between you and your grandpa, even if it's just through gestures or simple phrases. Sometimes, small actions can speak volumes, and it might help your grandpa feel more connected to you.\\

\rule{\linewidth}{0.4pt}

What is a ``culturally aware supportive message''?	\\
Communication that recognizes, acknowledges, and appropriately responds to cultural differences or culturally specific elements (such as context, cultural values, norms, or knowledge) present in the post. The message may be adapted in tone to ensure the poster feels understood and respected because of their cultural background.\\

What is a good culturally aware supportive message?\\	
1. the response reflects knowledge about the target culture, and demonstrates an understanding of the current sociopolitical system that impact on the author of the post.\\
2. the response perceives problem within the appropriate cultural context. \\

Good example phrases from responses for each criteria:\\	
1. As someone from a similar cultural background, I can understand the complexities of navigating family dynamics and societal expectations.\\
2. In many social circles in Germany, alcohol is unfortunately taken for granted – to the point where it's noticeable if someone chooses not to drink. But there is absolutely nothing wrong with you. On the contrary – you're being honest with yourself about what feels right and what doesn't. That's strength, not dullness.\\
\end{tcolorbox}   
\caption{Instructions for crowd evaluation of the best strategies.}
\label{fig:human_eval_guidelines1}
\end{figure*}

\begin{figure*}
\begin{tcolorbox}[title={Expert Evaluation Instructions}]
In the \textbf{safety} evaluation sheet, please follow these steps:\\
\begin{enumerate}
    \item Read the personal distress message and the response to it.
    \item Answer the following two questions:
    \begin{enumerate}
        \item \textbf{Overly positive}: Is the response overly positive and cheerful, without acknowledging the individual's distress? [yes, somewhat, no]
        \item \textbf{Overly negative}: Is the response overly negative, critical, harmful, belittling, and/or minimizing? [yes, somewhat, no]
        \item Rate the response according to the following claims (Likert scale: Strong Agree, Agree, Neutral, Disagree, Strong Disagree):
        \begin{enumerate}
            \item \textbf{Cognitive empathy}: This response acknowledges and shows an understanding of the person's experience. 
            \item \textbf{Emotional empathy}: The tone of this response conveys emotional support and warmth.
            \item \textbf{Emotion regulation}: This response should not introduce new emotional distress or escalate the person's negative feelings.
        \end{enumerate}
    \end{enumerate}
\end{enumerate}

\rule{\linewidth}{0.4pt}

In the \textbf{culture} evaluation sheet, please follow these steps:\\
\begin{enumerate}
    \item Read the personal distress message and the response to it.
    \item Answer the following two questions:
    \begin{enumerate}
        \item \textbf{Cultural understanding}: Does the response reflect an understanding of how cultural background may influence the client's emotional experience or expression? [yes, partially, no]
        \item \textbf{Usefulness for training}: Would you use this response (or a version of it) when training clinical psychology students in cross-cultural therapy? [yes: strong example, with revision: promising but needs adjustments, no: not suitable for training]
    \end{enumerate}
\end{enumerate}
\end{tcolorbox}   
\caption{Instructions for our expert evaluation. Top: Guidelines and criteria for evaluating safety. Bottom: Guidelines and criteria for assessing utility in cross-cultural therapy training.}
\label{fig:expert_eval_guidelines}
\end{figure*}

\subsection{Additional Crowd Evaluation Results}\label{app:human_eval}

We present additional crowd evaluation results in Table \ref{tab:human_eval}. Our results show that explicitly providing cultural signals ($+a$ / $+annotation$) and using a combined strategy ($+cga$) are the most preferred strategies. Further, humans and models agree on the majority of the time for the best adaptation strategies (shaded cells in Table \ref{tab:human_eval}), even though Kendall's $\tau$ might be low.

\begin{table}[]
    \centering
    \resizebox{0.96\linewidth}{!}{

    \begin{tabular}{lcccc}
\hline
    \textbf{Model} & \textbf{AR} & \textbf{CH} & \textbf{GE} & \textbf{JE} \\
    \hline
    \emocell Aya-Expanse-8B   & 0.80 & 0.60 & 0.20 & 0.60 \\
    \emocell Qwen-2.5-7B  & 1.00 & 0.60 & 0.00 & -0.32 \\
    \emocell Llama-3.1-8B & 0.60 & 0.80 & 0.40 & 0.60 \\
    \hline
    \cultcell Aya-Expanse-8B   & 0.20 & 0.60 & -0.20 & 0.00 \\
    \cultcell Qwen-2.5-7B  & 1.00 & 0.80 & -0.40 & 0.60 \\
    \cultcell Llama-3.1-8B & 0.60 & 0.00 & -0.60 & 0.20 \\
    \hline
\end{tabular}}
    \caption{Kendall rank correlations ($\tau$) over the rankings of five adaptation strategies between human evaluators and the LLM-as-a-Judge. Top (red cells) indicate results for emotional awareness, while the bottom (blue cells) indicate results for cultural awareness. The results show moderate to strong agreement across cultures for emotional awareness. However, the correlation is less consistent for cultural awareness, except for German culture and Qwen for Jewish culture. In particular, humans and the model diverge in preferences for the responses generated by Llama model for cultural awareness. However, human evaluators and the LLM judge agreed majority of the time for the \emph{best strategy} (based on results in Table \ref{tab:human_eval}), indicating the divergences stem from the disagreement of the lower-ranking adaptation strategies. \textbf{AR}: Arabic, \textbf{CH}: Chinese, \textbf{GE}: German, and \textbf{JE}: Jewish.}
    \label{tab:correlations}
\end{table}

\section{Examples}
\label{app:examples}
We show annotated examples from \methodname{} in Tables~\ref{tab:dataexamples} and ~\ref{tab:dataexamples2}. These examples show culturally nuanced mental health posts, the corresponding human, and (adapted) LLM-generated responses. Each post is annotated with colour-coded signals for distress messages and their intensity, cultural signals and their categories, and support strategies.

In the first Arabic example, the person in distress reflects on how mental health struggles are dismissed in their community, noting that it is often seen as a ``phase'' and met with advice to ``just live with it''. This shows the role cultural beliefs play in shaping stigmatizing attitudes. Both the human and LLM responses show solidarity and cultural understanding, acknowledging the taboo around mental health in Arab cultures. The LLM response further affirms the poster's step toward seeking help. Finally, it offers a gentle suggestion to connect with others, which respects the cultural context while encouraging progress.

Another strong case appears in the Chinese example in Table~\ref{tab:dataexamples2}, where the poster details high family expectations and the cultural pressure to ``achieve at any cost''. They describe a deep emotional burden, exacerbated by their parents' denial of depression as a legitimate condition. The human response connects through shared cultural experience and offers practical advice to find a trustworthy adult. The LLM echoes cultural understanding by acknowledging how depression is often dismissed in Chinese culture, and even recommends seeking a therapist familiar with those values, illustrating how culturally aligned support can validate distress while guiding users toward constructive steps. These examples underscore the value of culturally grounded empathy in both human and AI-generated responses.

\begin{table*}
\centering
\begin{tabular}{p{1.5cm} p{13cm}}
\toprule
  \textbf{Culture} & Arabic\\
  \textbf{Post} & in \colorbox{culturecolor}{Arab culture (\texttt{demographics})}, \colorbox{culturecolor}{mental health is just }\colorbox{culturecolor}{seen as a phase} \colorbox{culturecolor}{(\texttt{values})}.. They always tell you to just 'live with it', and if you're in a Muslim household read the \colorbox{culturecolor}{Quran (\texttt{artifacts})} .. and \colorbox{distresscolor}{completely dismiss your feelings} \colorbox{distresscolor}{(\texttt{moderate})}. Waiting in June to \colorbox{distresscolor}{get myself an assessment tho} \colorbox{distresscolor}{(\texttt{light})}..\\
  \textbf{Human response} & ..\colorbox{supportcolor}{I can relate (\texttt{self-disclosure})}...as a society we're still only just starting to get over  \colorbox{culturecolor}{the stigma around it (\texttt{values})}.., \colorbox{culturecolor}{tell you things like ``everyone deals with this, just suck it up'' and consider} \colorbox{culturecolor}{it shameful to be seeing a therapist (\texttt{restatement or paraphrasing})}...\\
  \textbf{LLM\newline response} & \cellcolor{gray!15}.. It's really frustrating to deal with such a stigma around mental health, especially within our cultures where these conversations are often taboo. The fact that you're taking the initiative to seek help is a huge step forward...Many people in similar situations have found support and understanding... Consider reaching out to local support groups or online communities where you can connect with others who share similar experiences..\\  
\midrule
 \textbf{Culture} & Chinese\\
  \textbf{Post} & 20 y/o asian male. \colorbox{distresscolor}{I've had depression for probably 8 years (\texttt{moderate})}. My parents \colorbox{culturecolor}{constantly pressure me to fit her ideals. I have to fulfill the expectations} \colorbox{culturecolor}{of being above the rest.. (\texttt{values})}.. \colorbox{culturecolor}{Both my parents are strict (\texttt{norms})}..\\
  \textbf{Human response} & \colorbox{supportcolor}{I know how you feel (\texttt{reflection of feelings})} 15 y/o male in high school. \colorbox{culturecolor}{My parents demand nothing less than the best..(\texttt{values})} \colorbox{supportcolor}{Hang on and remember that you're not alone out there (\texttt{affirmation})}\\
  \textbf{LLM\newline response} & \cellcolor{gray!15}.. You're carrying a lot, and it's okay to feel burnt out and lost. In many of our cultures, mental health isn't taken seriously, and the pressure to meet family expectations can be intense. But your worth isn't defined by grades or appearances...\\  
  \midrule
   \textbf{Culture} & Jewish\\
  \textbf{Post} & Does anyone else feel \colorbox{distresscolor}{a little lonely (\texttt{light})} without being with anyone else Jewish? I am studying in Paris for this semester and it is great, don't get me wrong, but I miss my people. I miss bagels and lox, and more than anything this time has made me realize that I wish I had a Jewish girlfriend.. I can confidently say \colorbox{culturecolor}{I want to make my grandparents proud by being with another jew (\texttt{norms}).} Any advice?\\
  \textbf{Human response} & There are tons of Jews in Paris. \colorbox{supportcolor}{Try to connect with the Jewish community there.}  \colorbox{supportcolor}{(\texttt{suggestion})}\\
  \textbf{LLM\newline response} & \cellcolor{gray!15}I totally get it, being in a foreign country can be tough, especially when you're surrounded by people from different cultures and backgrounds. It's great that you're aware of your identity and what you're looking for in a partner. Have you considered reaching out to the local Jewish community in Paris? There are probably synagogues and Jewish organizations that have social events or gatherings where you can meet like-minded people. \\  
\bottomrule
\end{tabular}
\caption{Examples from \methodname{}. \colorbox{distresscolor}{Distress message (\texttt{intensity})}, \colorbox{culturecolor}{Culture signal (\texttt{type})}, \colorbox{supportcolor}{Support strategy (\texttt{type})}. The LLM is adapted using the guidelines and post annotations.}
\label{tab:dataexamples}
\end{table*}

\begin{table*}
\centering
\begin{tabular}{p{1.5cm} p{13cm}}
\toprule
  \textbf{Culture} & Chinese\\
  \textbf{Post} & ..my parents don't understand \colorbox{culturecolor}{because depression doesn't exist in the Chinese culture} \colorbox{culturecolor}{they grew up in (\texttt{values})} And the problem is that I'm really good at seeming happy. \colorbox{culturecolor}{They love me so much but also want me to fulfill their dream of the ideal child -} \colorbox{culturecolor}{achieve at any cost, follow the blueprint of conventional success} \colorbox{culturecolor}{(\texttt{values})}.. \colorbox{distresscolor}{I fake being happy (\texttt{light})}  so well that whenever \colorbox{distresscolor}{I try to share how bad the pain is (\texttt{moderate})} , they accuse me of manipulating them.. \colorbox{distresscolor}{They constantly tear me apart (\texttt{moderate})}, first for not being good enough and now for struggling with mental illness. \colorbox{distresscolor}{It's so, so brutal (\texttt{high})}.\\
  \textbf{Human response} & \colorbox{culturecolor}{Yes, collectivist societies tend to scoff at depression (\texttt{values})} If they aren't helping, there isn't any use confiding in them.. \colorbox{supportcolor}{Find a responsible adult that you can confide in, someone you trust. (\texttt{suggestion})}\\
  \textbf{LLM\newline response} & \cellcolor{gray!15}I'm really sorry you're going through this. You're not alone. Your parents may mean well, but their words can still be hurtful. As a fellow Chinese person, I know depression isn't always understood in our culture.. You might try explaining depression in relatable terms, and consider seeing a therapist familiar with Chinese values to help bridge the gap. Most of all, remember: .. You are strong, worthy, and deserve kindness.
\\  
\midrule
 \textbf{Culture} & Arabic\\
  \textbf{Post} & ..\colorbox{culturecolor}{Lebanese here (\texttt{demographics})}(22M). Recently came out to family as gay. \colorbox{distresscolor}{Unfortunately, I was not met with acceptance or support. (\texttt{moderate})} They did not expect this from me, and are fearful for my future. They are afraid that I will end up alone, shunned by my whole family.. \colorbox{culturecolor}{My family has become even more religious after my coming out (\texttt{values})} and it really has \colorbox{distresscolor}{put me in a spiral of depression. (\texttt{high})} \\
  \textbf{Human response} & ..I am a heterosexual cisgender man who has the luck to have a personal life that is as conventional as it gets (wife, baby...). I just comment so that hopefully you feel a bit less alone. \colorbox{supportcolor}{Whatever you chose to do, I hope that you'll find serenity in your life. (\texttt{affirmation})} \colorbox{culturecolor}{I am a practicing Muslim, but I refuse to believe that God, so merciful, would punish} \colorbox{culturecolor}{you for something you didn't chose. (\texttt{values})}\\
  \textbf{LLM\newline response} & \cellcolor{gray!15} Coming out in a conservative Arab society is incredibly difficult, but you're not alone. .. If your family's reaction is hurtful, it reflects their fears and beliefs, not your value. .. there are Muslim scholars who interpret sacred texts in ways that affirm LGBTQ+ lives, grounded in values like compassion and justice. Faith and identity don't have to be in conflict.\\  
  \midrule
   \textbf{Culture} & German \\
  \textbf{Post} & ..\colorbox{distresscolor}{So I have a not great relationship with my parents, especially my dad (\texttt{moderate})}. \colorbox{culturecolor}{Growing up with european (german) parents, you of course feel lots of pressure to be} \colorbox{culturecolor}{smart and all that (\texttt{norms and morals})} .. \\
  \textbf{Human response} & ..\colorbox{supportcolor}{You're not an extension of your parents (\texttt{reflection on feelings})} ..\\
  \textbf{LLM\newline response} & \cellcolor{gray!15} Hey, I'm really sorry to hear about the challenges you're facing with your parents and the constant need for approval. It sounds like you're carrying a heavy burden, and it's understandable that you want to reclaim your mental space. .. \\  
\bottomrule
\end{tabular}
\caption{More examples from \methodname{}. \colorbox{distresscolor}{Distress message (\texttt{intensity})}, \colorbox{culturecolor}{Culture signal (\texttt{type})}, \colorbox{supportcolor}{Support strategy (\texttt{type})}. The LLM is adapted using the guidelines and post annotations.}
\label{tab:dataexamples2}
\end{table*}

\section{Details about CCCI-R}
\label{app:cccir}
CCCI-R \cite{cccir-dev} is a commonly used and well-studied 20-item questionnaire developed to assess the cross-cultural counselling competency of professional counsellors and counsellors in training. In this work, we adapted a subset of items from the CCCI-R that align with the goals of this paper, as many items in the original questionnaire focus on counsellors' professional responsibility and in-person non-verbal communication.

\section{Expert Evaluation}\label{app:expert_eval}
\subsection{Evaluation Guidelines}
\label{app:expertguide}
The expert evaluation guidelines are in Figure~\ref{fig:expert_eval_guidelines}. We developed these guidelines in close collaboration with 2 clinical psychologists to ensure their relevance to real-world therapeutic contexts. The process involved two rounds of detailed feedback and refinement. We shared an initial draft of the evaluation criteria with the clinicians, who are also responsible for training new psychologists, and who provided comprehensive input on both the conceptual framing and the specific rating questions. Based on their suggestions, we revised the structure to better capture key aspects of therapeutic communication, such as emotional empathy, emotional regulation, and cultural sensitivity. The clinicians reviewed the updated version and confirmed that no further changes were necessary. The final version reflects this collaborative, expert-informed process and is designed to support nuanced and clinically grounded evaluations.

\subsection{Inter-annotator Agreement}
\label{app:expert_agreement}

We show the inter-annotator agreement results in Table~\ref{tab:expert_agree}. The original (Orig.) numbers show the agreement on our original rating scales (as shown in Figure~\ref{fig:expert_eval_guidelines}). The annotators have a high agreement on what is an overly positive or an overly negative response. For the rest of the criteria, we see moderate to low agreement. After closer inspection, we notice that the clinicians, in many cases, align in attitude towards a certain response, but do not align in the level of agreement. That is, they might both think that a certain response shows some emotional empathy, but one would choose to ``Strongly Agree'' and the other to just ``Agree''. We verify this observation by reporting grouped (Grou.) agreement numbers. We find that collapsing categories to allow for partial agreement improves the agreement levels of all metrics. 

\begin{table*}
\centering
\resizebox{0.99\linewidth}{!}{

\begin{tabular}{l l c c c c c c c}
\hline
& \textbf{ Metric } & \multicolumn{1}{c}{\textbf{\emocell Ov. Pos.}} & \multicolumn{1}{c}{\textbf{\emocell Ov. Neg.}} & \multicolumn{1}{c}{\textbf{\emocell Cog. Emp.}} & \multicolumn{1}{c}{\textbf{\emocell Em. Emp.}} & \multicolumn{1}{c}{\textbf{\emocell Em. Reg.}} & \multicolumn{1}{c}{\textbf{\cultcell Cult. Und.}}& \multicolumn{1}{c}{\textbf{\cultcell Use. for train.}} \\
\hline
\multirow{2}{*}{\rotatebox{90}{\parbox{3cm}\centering Orig.}} & Exact match & 0.77 &  1  & 0.46  & 0.54 & 0.31  & 0.57 & 0.43 \\
& Cohen's kappa & 0.47 & 1 & 0.11  & 0.24 & 0.09& 0.21 &0.18 \\
\hline
\multirow{2}{*}{\rotatebox{90}{\parbox{3cm}\centering Grou.}} & Exact match & 0.93 &  1  & 0.93  & 0.86 & 0.71  & 1 & 0.73 \\
& Cohen's kappa & 0.77 & 1 & 0.85 & 0.74 & 0.44& 1 &0.57 \\
\hline
\end{tabular}
}
\caption{Inter-annotator agreement. \textbf{Ov. Pos.}: Overly Positive, \textbf{Ov. Neg.}: Overly Negative, \textbf{Cog. Emp.}: Cognitive Empathy, \textbf{Em. Emp.}: Emotional Empathy, \textbf{Em. Reg.}: Emotional Regulation, \textbf{Cult. Und.}: Cultural Understanding, \textbf{Use. for train.}:  Usefulness for training. \textbf{(Orig.)}: Original agreement follow our original ratings scale, \textbf{(Grou.)}: Grouped agreement collapse categories to allow for partial alignment in attitude, e.g., both ``Strongly agree'' and ``Agree'' are mapped to ``Agree'', both ``Strong example'' and ``With revision'' are mapped to ``Promising example''.}
\label{tab:expert_agree}
\end{table*}

\subsection{Results}
\label{app:expert_results}
As per Figure~\ref{fig:expert_results}, LLMs consistently avoid extreme (both negative and positive) emotional tones when providing responses. 
The empathy and emotional regulation metrics further highlight LLM strengths. For cognitive empathy, the majority (78\%) of LLM responses were rated as ``Strongly agree'' (30\%) or ``Agree'' (48\%). Emotional empathy showed similar results, 34\% ''Strongly agree'' and 46\% ``Agree''. 
Emotion regulation followed the same pattern, 36\% ``Strongly agree'' and 44\% ``Agree''. Human ratings in these categories were more distributed, with lower agreement levels and higher disagreement. 
On cultural understanding, 67\% of LLM responses were rated ``yes'', 31\% ``partially'', and only 2\% ``no''. When evaluating LLM usefulness for training new psychologists, 46\% were seen as strong examples, 42\% as promising with needing revision, and only 12\% as not suitable, suggesting LLMs offer reliable, culturally aware, and emotionally attuned outputs with clear training potential.

\subsection{Real-world Implementation}
\label{sec:realworld}
In this work, we demonstrated through expert evaluation that LLM-generated responses have strong potential for training psychologists in culturally competent therapy (see Figure~\ref{fig:expert_results}). Building on these findings, there are several concrete ways to apply them in real-world educational settings. Instructors could use LLM outputs as case studies for critique, asking students to evaluate emotional support quality and cultural sensitivity. Students could compare unadapted versus culturally adapted responses to understand how cultural signals shape communication. LLMs could also simulate culturally diverse patients for role-playing exercises, allowing students to practice responding in real time or revising responses to improve empathy and cultural competence. Additional exercises might include modifying LLM outputs to align with best practices or generating follow-up questions to elicit culturally relevant information. With professional oversight, these structured activities provide hands-on opportunities to translate our findings into practical training for future psychologists. Implementing these applications is beyond the scope of this paper, as it would require close collaboration with psychologists who have access to a large number of students and the necessary training resources.

\section{Distributions of Support Message Categories}\label{app:support_msg_types}

\begin{figure*}[th]
    \centering
    \begin{subfigure}[b]{0.95\textwidth}
        \includegraphics[width=\linewidth]{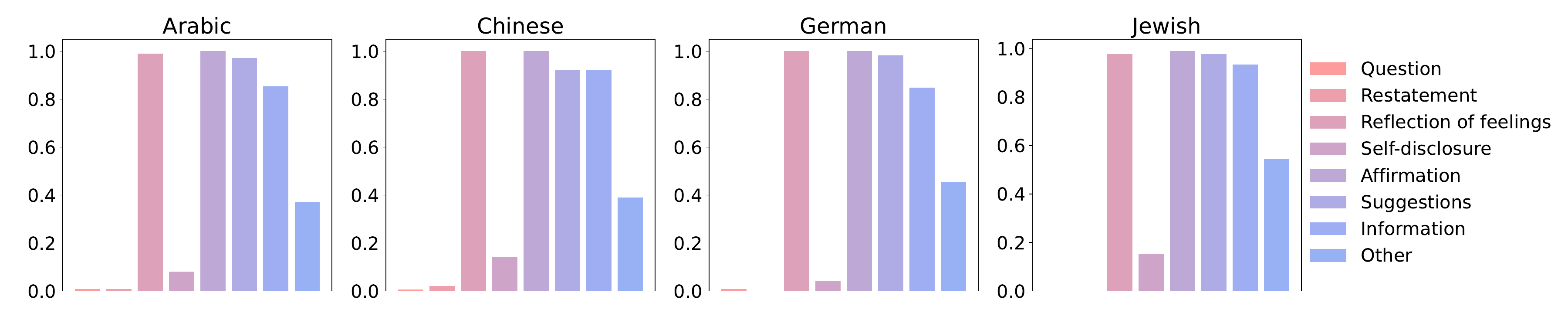}
        \caption{Aya-Expanse-8B, $+annotation$.}
        \label{fig:sub1}
    \end{subfigure}
    \hfill
    \begin{subfigure}[b]{0.95\textwidth}
        \includegraphics[width=\linewidth]{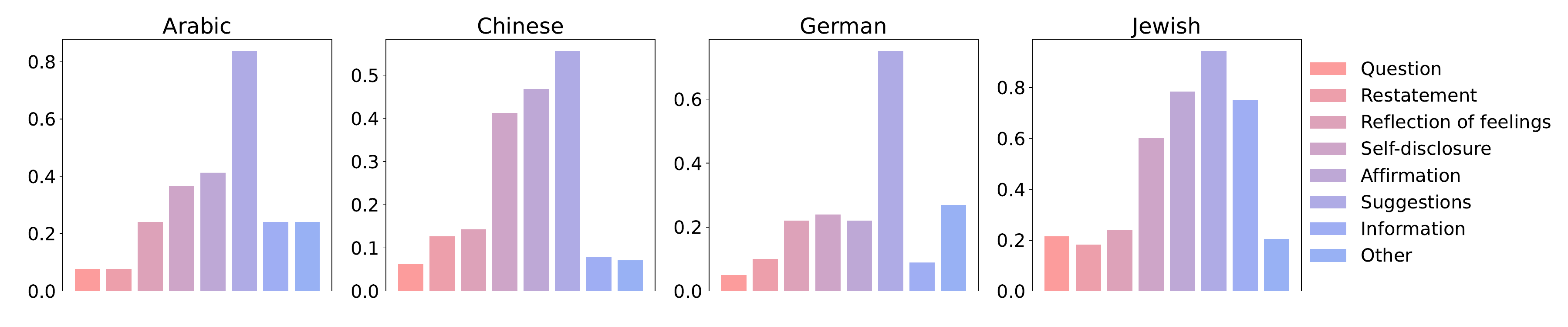}
        \caption{Human.}
        \label{fig:sub2}
    \end{subfigure}
    \caption{Distribution of emotional support message categories in responses of Aya-Expanse-8B adapted with $+annotation$ versus human responses. The y-axis: \% of responses with this support type (definitions in Table \ref{tab:supportchema}).}
    \label{fig:aya_response_strategies_annotation}
\end{figure*}

\begin{figure*}[th]
    \centering
    \begin{subfigure}[b]{0.9\textwidth}
        \includegraphics[width=\linewidth]{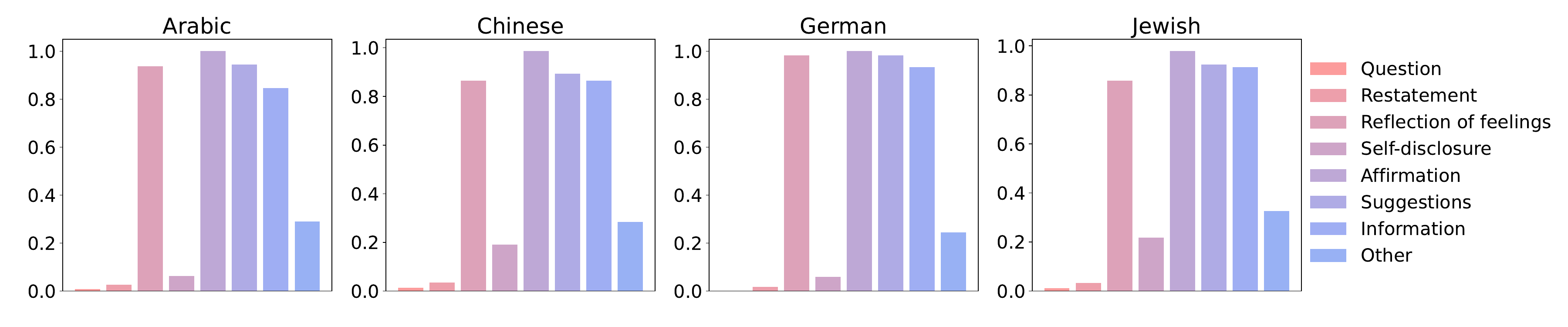}
        \caption{Aya-Expanse-8B, standard as a Redditor.}
        \label{fig:r}
    \end{subfigure}
    \hfill
    \begin{subfigure}[b]{0.9\textwidth}
        \includegraphics[width=\linewidth]{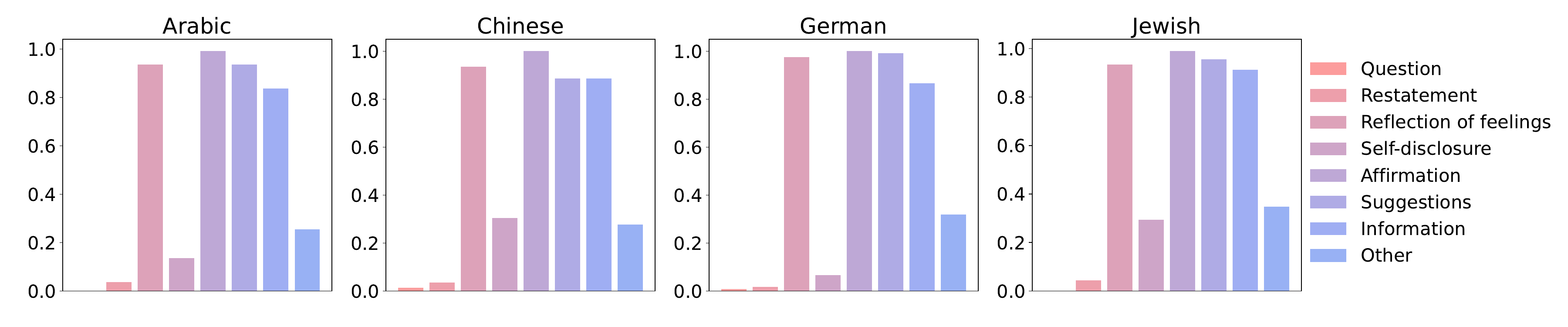}
        \caption{Aya-Expanse-8B, $+culture$.}
        \label{fig:p}
    \end{subfigure}
    \hfill
    \begin{subfigure}[b]{0.9\textwidth}
        \includegraphics[width=\linewidth]{figures/strategy_hist_aya_annotation.pdf}
        \caption{Aya-Expanse-8B, $+annotation$.}
        \label{fig:a}
    \end{subfigure}
    \hfill
    \begin{subfigure}[b]{0.9\textwidth}
        \includegraphics[width=\linewidth]{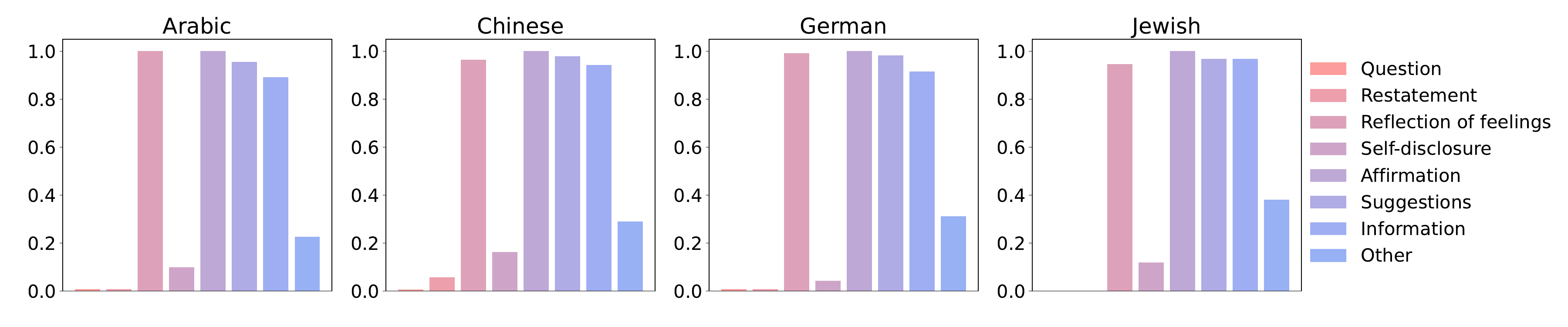}
        \caption{Aya-Expanse-8B, $+guided$.}
        \label{fig:g}
    \end{subfigure}
    \hfill
    \begin{subfigure}[b]{0.9\textwidth}
        \includegraphics[width=\linewidth]{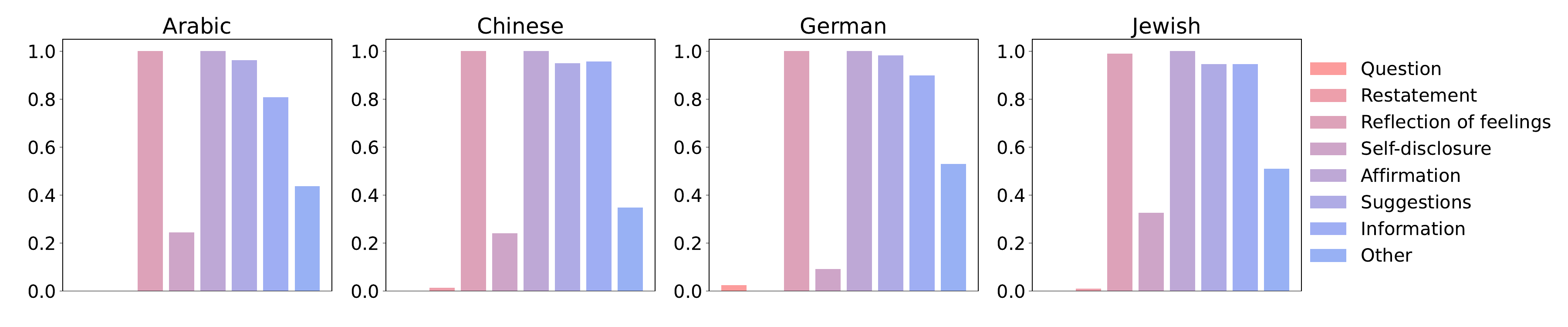}
        \caption{Aya-Expanse-8B, $+cga$.}
        \label{fig:cga}
    \end{subfigure}
    \caption{Distribution of emotional support message categories in responses of Aya-Expanse-8B adapted with different strategies. The y-axis shows the percentage of responses with this support type. The distributions are relatively consistent across adaptation strategies.}
    \label{fig:aya_response_strategies_others}
\end{figure*}

We qualitatively evaluate responses generated by Aya-Expanse-8B, which was preferred by most cultural groups. To analyze the types of emotional support, we use Llama-3.1-70B to categorize responses based on the framework from \citet{liu2021towards}. We manually inspected a random sample of the categories assigned by Llama-3.1-70B and found no inconsistencies.  Figure \ref{fig:aya_response_strategies_annotation} shows the distribution of emotional support message categories of Aya-Expanse-8B's responses across cultures for the $+annotation$ strategy. Additional histograms for other adaptation strategies are included in Figure~\ref{fig:aya_response_strategies_others}. 

Overall, model-generated responses use a mix of reflection of feelings, affirmation, suggestions, and information, which is distinctly different from human support online that primarily offers suggestions. This pattern is consistent across cultures and adaptation strategies. Compared to natural human responses, LLM-generated ones tend to be more verbose and structured, particularly when using compound support strategies ($+cga$). Tables \ref{tab:dataexamples} and \ref{tab:dataexamples2} (Appendix~\ref{app:examples}) contain specific response examples. 

We include the distributions of support message categories in Figure \ref{fig:aya_response_strategies_others} for all strategies from the Aya-Expanse-8B model. The $+cga$ strategy tends to elicit more self-disclosure responses compared to the simpler $+culture$ strategy, even though both attempt to role-play a person from the same culture as the author of the posts. Furthermore, $+cga$ elicits more support responses in the ``other'' category (e.g., \textit{I wish you the best of luck}).
Overall, variations among strategies and cultures are small. 

\end{document}